\documentclass[sigconf, screen]{acmart}

\usepackage{multirow}
\usepackage{booktabs}
\usepackage{bbding}
\usepackage{pifont}
\usepackage{utfsym}
\usepackage{makecell}
\usepackage{array}
\usepackage{amsmath}
\usepackage[inline]{enumitem}
\usepackage{graphicx}
\usepackage{adjustbox}
\usepackage{csquotes}
\usepackage{newfloat}
\usepackage{listings}
\usepackage{threeparttable}

\AtBeginDocument{%
  }

\setcopyright{acmlicensed}
\copyrightyear{2025}
\acmYear{2025}
\acmDOI{XXXXXXX.XXXXXXX} 
\acmISBN{978-1-4503-XXXX-X/2018/06} 
\acmConference[MM ’25]{33rd ACM International Conference on Multimedia}{October 27–31, 2025}{Dublin, Ireland}
\acmBooktitle{Proceedings of the 33rd ACM International Conference on Multimedia (MM ’25), October 27–31, 2025, Dublin, Ireland}
\acmPrice{15.00}


\begin{document}

\title{MS-DETR: Towards Effective Video Moment Retrieval and
Highlight Detection by Joint Motion-Semantic Learning}

\author{Hongxu Ma}
\authornote{Both authors contributed equally to this research.}
\authornote{Work done during the internship at Tencent YouTu Lab.}
\affiliation{
  \institution{Fudan University}
  \city{Shanghai}
  \country{China}}
\email{hxma24@m.fudan.edu.cn}

\author{Guanshuo Wang}
\authornotemark[1]
\affiliation{
  \institution{Tencent Youtu Lab}
  \city{Shanghai}
  \country{China}}
\email{mediswang@tencent.com}

\author{Fufu Yu}
\affiliation{
  \institution{Tencent Youtu Lab}
  \city{Shanghai}
  \country{China}}
\email{fufuyu@tencent.com}

\author{Qiong Jia}
\affiliation{
  \institution{Tencent Youtu Lab}
  \city{Shanghai}
  \country{China}}
\email{boajia@tencent.com}

\author{Shouhong Ding}
\authornote{Corresponding author.}
\affiliation{
  \institution{Tencent Youtu Lab}
  \city{Shanghai}
  \country{China}}
\email{ericshding@tencent.com}

\begin{abstract}
Video Moment Retrieval (MR) and Highlight Detection (HD) aim to pinpoint specific moments and assess clip-wise relevance based on the text query. 
While DETR-based joint frameworks have made significant strides, there remains untapped potential in harnessing the intricate relationships between temporal motion and spatial semantics within video content.
In this paper, we propose the \textbf{M}otion-\textbf{S}emantics DETR (MS-DETR), a framework that 
captures rich motion-semantics features through unified learning for MR/HD tasks. 
The encoder first explicitly models disentangled intra-modal correlations within motion and semantics dimensions, guided by the given text queries.
Subsequently, the decoder utilizes the task-wise correlation across temporal motion and spatial semantics dimensions to enable precise query-guided localization for MR and refined highlight boundary delineation for HD.
Furthermore, we observe the inherent sparsity dilemma within the motion and semantics dimensions of MR/HD datasets.
To address this issue, we enrich the corpus from both dimensions by generation strategies and propose contrastive denoising learning to ensure the above components learn robustly and effectively.
Extensive experiments on four MR/HD benchmarks demonstrate that our method outperforms existing state-of-the-art models by a margin. Our code is available at \url{https://github.com/snailma0229/MS-DETR.git}.
\end{abstract}


\maketitle

\section{Introduction}
In recent years, online video content has a remarkable surge~\cite{ma2025generativeregressionbasedwatch}, attributable to its diverse and easily accessible nature. Compared with images and text, videos offer a wealth of information across various modalities. However, this abundance of information also inevitably extends the time needed to browse and pinpoint content of interest. Consequently, relevant clips localization tasks such as Moment Retrieval (MR) and Highlight Detection (HD) have emerged.
The MR task aims to retrieve the specific moments most relevant to a given language description, and predict the accurate boundaries.
Similar but different, the HD task predicts the relevance of all given clips to the text descriptions for salience assessment.
Considering the task similarity between them, \cite{lei2021detecting} firstly proposed the joint learning of MR and HD tasks within a unified framework based on DETR~\cite{carion2020end}, which is inherited by most subsequent works~\cite{lei2021detecting,liu2022umt,lin2023univtg,moon2023query,jang2023knowing,xu2023mh,sun2024tr} for joint MR/HD tasks.

\begin{figure*}[t]
  \centering
   \includegraphics[scale=0.78]{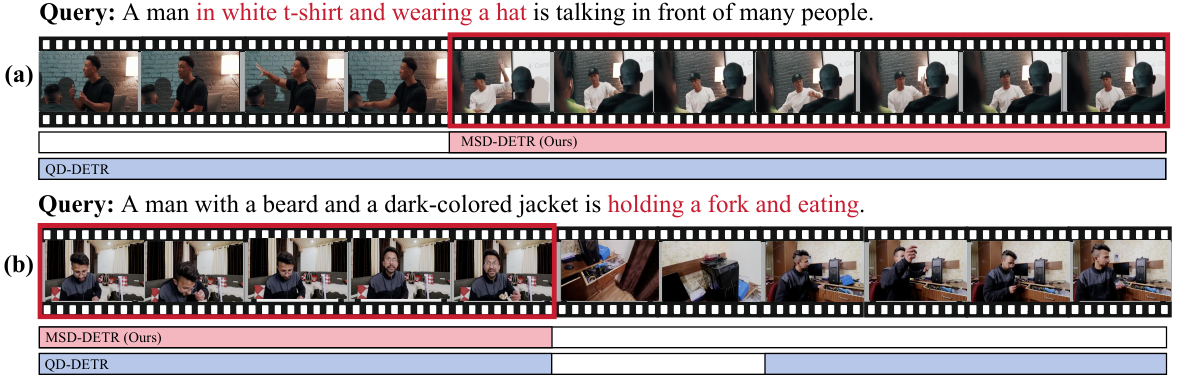}
   \caption{Examples from the QVHighlight dataset~\cite{lei2021detecting} illustrating our motivation: the performance by MR/HD may be determined by fine-grained temporal motion and spatial semantic clues within video-text pairs, so more accurate results require learning finer-grained motion-semantics representations and effectively utilizing deeper intrinsic associations between them. (Best viewed zoomed in on screen)}
   \label{fig:motivation}
\end{figure*}

Dynamic temporal concepts in video contents like motion and visual semantics information like scene features play a crucial role in isolating a specific moment from the others within a video.
For example, in Fig.~\ref{fig:motivation}(a), both of the two men are talking in front of the audience, demonstrating similar temporal motion cues, we can only accurately distinguish moments by their appearance, \textit{i.e. spatial semantics features}. 
In Fig.~\ref{fig:motivation}(b), 
the same individual with a beard and dark jacket appears at various points through the video, thus understanding the action described in text query, \textit{i.e. temporal motion features}, is essential to recognize moments of interest.
These examples show that performance by MR/HD may be determined by \textit{fine-grained temporal motion and spatial semantic clues} within video-text pairs, so more accurate results require learning finer-grained motion-semantics representations and effectively utilizing deeper intrinsic associations between them, which are yet widely overlooked as follows:

First, video features are represented by pre-trained motion and semantic components~\cite{radford2021learning, feichtenhofer2019slowfast} in joint MR/HD tasks. But existing works tend to ignore the intrinsic discrimination within both temporal- and spatial-biased features, and just concatenate them as global video representations.
The global video representations are utilized to concatenate with text query embeddings~\cite{lei2021detecting} or conduct cross-modal attention with text query embeddings~\cite{moon2023query,xu2023mh} as a unified input fed into the encoder, which may restrict the fine-grained discrimination in the video contents.

Second, the typical DETR-like encoder-decoder framework in existing works~\cite{lei2021detecting,liu2022umt,moon2023query,jang2023knowing,xu2023mh,sun2024tr} process the MR/HD tasks separately, but the intrinsic relationship of motion-semantics dimensions between both tasks are also neglected. 
Specifically, HD performed by encoder imposes influence on MR solely through encoded embeddings~(memory), but MR has almost rare direct impact on HD, which reflects a potential information bottleneck. 
According to the given text query, semantics salience is predicted along the temporal motions dimension in HD task, which could be a prior preference guidance for MR. On the contrary, the precise temporal boundaries and semantic foregrounds in MR task could supplement accurate constraints beside subjective ratings in HD.

Lastly, a fundamental challenge is still hard to address: the sparsity dilemma of temporal motion and spatial semantics dimensions inherent in MR/HD datasets annotations, \textit{i.e.}, the semantics richness of both dimension within a video far exceeds that of a given limited-length text description, which renders the joint MR/HD frameworks that use a single text as the query incapable of generalizing effectively to accurately discriminate fine-grained video contents.

In this paper, we propose the Motion-Semantics Detection Transformer (MS-DETR), a novel joint MR/HD framework, which advances motion and semantic joint learning strategy as follows:
Towards comprehensive query-based video representations, we first design a Motion-Semantics Disentangled Encoder (MSDE), which explicitly distinguishes the temporal motion and spatial semantics dimensions within videos, and interacts more refined information with text query in both respective dimensions.
Then, to facilitate better synchronization and synergy between the MR and HD tasks of motion-semantics dimensions, we propose the Mutual Task-Collaborated Decoder (MTCD). We utilize the salience predictions by the HD task to dynamically generate temporal position queries with motion and content queries with prior semantics. The queries by HD serve as a prior preference guidance to localize relevant moments in the MR task.
In a reciprocal manner, accurate temporal boundaries and semantic foreground constraints in the MR task are utilized to refine the discrimination of salience prediction in the HD task.
Last, to address the inherent sparsity annotation dilemma problems, we enrich the data from both temporal motion and spatial semantics dimensions, aiming to achieve richness alignment between video and text at the data level in terms of motion-semantics dimensions. 
To deal with the inevitable noise within auxiliary data for full utilization, we introduce the denoising training strategy for reliable training and conduct explicit contrastive metric learning between matched (pos) and unmatched (neg) video-text pairs.

Our contributions are as follows:
\begin{enumerate*}[label=(\roman*)]
    \item We propose a unified MR/HD framework MS-DETR, consisting of MSDE to learn more nuanced motion-semantics disentangled video representations based on text query, and MTCD to utilize the mutual synergistic benefits between MR/HD in motion-semantics dimensions.
    \item We introduce the corpus generation to address the sparsity annotation dilemma of both dimensions and propose contrastive denoising learning for robust learning from generated data.
    \item Our method outperforms all existing SOTAs by a remarkable margin over four benchmarks.
\end{enumerate*}

\section{Related Work}
\subsection{Moment Retrieval and Highlight Detection}

Moment retrieval (MR) aims to retrieve video moments based on text queries. Proposal-based MR methods generate candidate moments based on the sliding window~\cite{gao2017tall,ge2019mac,liu2018cross} or proposal generation networks~\cite{chen2018temporally,xu2019multilevel,yuan2019semantic,zhang2019man,zhang2019cross} and then select high-scoring candidates based on similarity to the query. These methods often involve complex pre-processing and post-processing steps, which can be suboptimal.  Proposal-free methods~\cite{ghosh2019excl,zhang2021natural,mun2020local} directly predict the start and end times of moments but generally achieve lower accuracy.
Different from MR, highlight detection (HD) originally focused on identifying key video clips without queries, evolving to include user preferences via textual queries~\cite{kudi2017words}. Existing HD methods can be categorized into supervised~\cite{gygli2016video2gif,sun2014ranking,xu2021cross}, weakly supervised~\cite{cai2018weakly,panda2017weakly,xiong2019less}, and unsupervised groups~\cite{badamdorj2022contrastive,khosla2013large,mahasseni2017unsupervised,rochan2018video}.

Despite the substantial correlation between the two tasks, they have not been jointly studied until QVHighlights~\cite{lei2021detecting}, which also proposed a baseline called Moment-DETR for jointly MR/HD. Most of the following works~\cite{moon2023query,xu2023mh,jang2023knowing,moon2023correlation,xiao2023bridging,sun2024tr} are based on this DETR-like framework. UMT~\cite{liu2022umt} exploited additional audio modality to achieve a unified multi-modal framework and perform flexible MR/HD. UniVTG~\cite{lin2023univtg} and Unloc~\cite{yan2023unloc} unified the diverse tasks architecture and emphasized the significance of pre-training. Different from previous methods, this paper focuses on disentangled and joint motion-semantics learning to effectively accomplish joint MR/HD.

\subsection{Vision-text Multi-modal Alignment}

Understanding video content fundamentally relies on aligning visual and textual modalities within joint embedding spaces~\cite{zhang2024musetalk,zhang2024learning}. 
Research like CLIP~\cite{radford2021learning} has demonstrated the applicability of these techniques across various tasks. However, pre-trained models often face a domain gap when applied to specific tasks, prompting research into improving cross-modal interactions in downstream tasks to address these limitation~\cite{jia2021scaling,li2022blip,hong2020mini,ye2021temporal,xu2024enhancing,xu2025hunyuanportrait,zhang2024pixelfade,li2025vista,li2025visuals,lu2024coarse,mi2025data}.
Early work such as MINI-Net~\cite{hong2020mini}, simply concatenated feature vectors from both modalities. QD-DETR~\cite{moon2023query} utilizes negative-pair to enhance the model’s capability of exploiting the query information.
Despite advancements, most approaches fail to adequately capture the associations between temporal motion and spatial semantics within videos, often resorting to straightforward feature concatenation, which limits the capability of model to precisely retrieve fine-grained motion and semantics features.

\section{Method}
\label{method}

\begin{figure*}[t]
  \centering
   \includegraphics[scale=0.46]{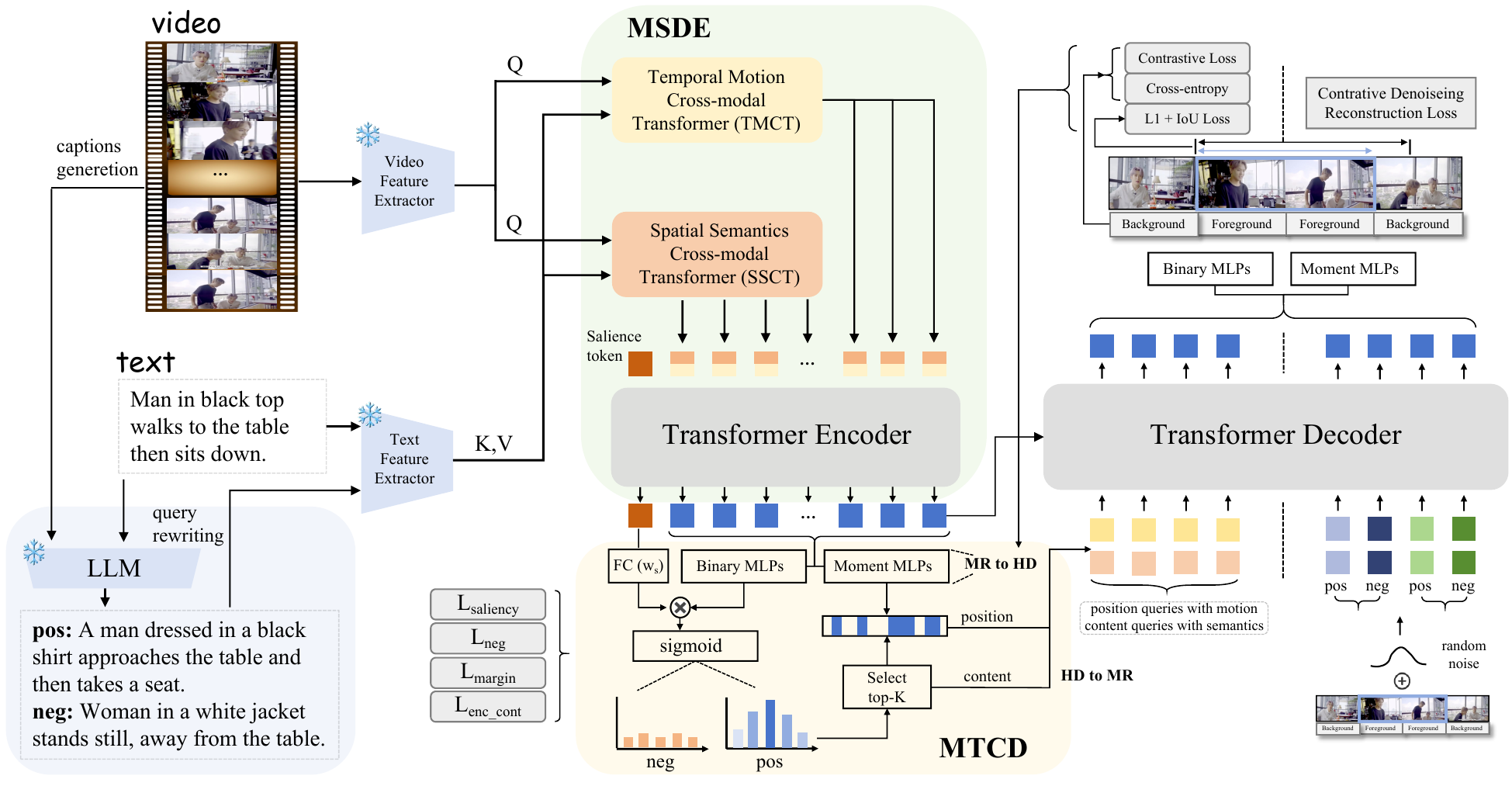}
   \caption{Overview of the MS-DETR's architecture, which consists of three core modules: (a) Motion-Semantics Disentangled Encoder (MSDE) explicitly distinguishes the temporal motion and spatial semantics dimensions within videos, and interacts more refined information with text query in both respective dimensions. (b) Mutual Task-Collaborated Decoder (MTCD) 
   utilizes the mutual synergistic benefits between MR/HD in motion-semantics dimensions and (c) Motion-Semantics Corpus Generation enriches the data from both temporal motion and spatial semantics dimensions to achieve richness alignment between video and text at the data level and Contrastive Denoising Training to ensure reliable training with the auxiliary data. Video clips in this figure are sampled from the QVHighlight dataset~\cite{lei2021detecting}. } 
   \label{fig:framework}
\end{figure*}

\subsection{Problem Formulation}
\label{problem_define}
Given an untrimmed video { $V$} of { $L$} clips and a text query { $Q$} with { $M$} words, clip-level video semantics features { $F^s_v = \{v^s_1, v^s_2, \ldots, v^s_L\} \in \mathbb{R}^{L \times d}$} and word-level text features { $F_t = \{t_1, t_2, \ldots, t_M\} \in \mathbb{R}^{M \times d} $} are extracted with frozen pre-trained CLIP~\cite{radford2021learning}. Clip-level video temporal motion features { $F^m_v = \{v^m_1, v^m_2, \ldots, v^m_L\} \in \mathbb{R}^{L \times d} $} are extracted with frozen SlowFast~\cite{feichtenhofer2019slowfast}. 
MR aims to retrieve the video moments that best match to the text query \textit{i.e.} the center and span { $\{c, s\}$} of each segment, and HD aims to predict salience scores { $S(\cdot)$} for all video clips. Fig. \ref{fig:framework} shows the overview of MS-DETR, consisting of our proposed core modules.

\subsection{Motion-Semantics Disentangled Encoder}

To effectively incorporate explicit contexts of both motion and semantics dimensions into query-based video representations, instead of concatenating into coupling video features, we introduce the Temporal Motion and Spatial Semantics Cross-modal Transformers (TMCT \& SSCT) to process interactions across different feature dimensions. Both towers consist of two layers of cross-attention transformers, and the \textit{key} {${K_t}$} and \textit{value} {${V_t}$} are both the query text features {${F_t}$}.
Corresponding \textit{queries} are respectively denoted as {$Q^m_v = F^m_v $} and {$Q^s_v = F^s_v $}. The query-based video representations {$\hat{F_v} = \{\hat{v_1}, \hat{v_2}, \ldots, \hat{v_L}\} \in \mathbb{R}^{L \times d}$} can be computed by:
\begin{equation}
    \hat{F_v} = \phi\left(\left(\mbox{TMCT}(Q^m_v, K_t, V_t)\oplus \mbox{SSCT}(Q^t_v, K_t, V_t)\right)\right)
  \label{F_v}
\end{equation}
where $\oplus$ refers to dimension-along concatenation, and $\phi(\cdot)$ is the learnable weighted outputs mapping to the same dimension $d$.
SSCT focuses on static visual details for contextual relevance within candidate clips. Conversely, TMCT captures the dynamics and transitions between consecutive clips for better sequence and duration understanding. 
The disentangling transformers enhance the perception of motion and semantic components of video content respectively, leading to more nuanced video representations.

Lastly, we concatenate {$\hat{F_v}$} with an salience token ${v_s}$, a randomly initialized learnable vector, to form {$\tilde{F_v} = \{{v_s},\hat{v_1}, \hat{v_2}, \ldots, \hat{v_L}\}$}. 
It serves as an input-adaptive predictor which interacts with query-based video representations {$\hat{F_v}$} as detailed in Eq.~(\ref{Salience}).
Then {$\tilde{F_v}$} are input into the Transformer Encoder to derive the output features $X=\{x_s, x_1, x_2, \ldots, x_L\}$, serving as the output for the HD task and as memory input for the decoder.

\subsection{Mutual Task-Collaborated Decoder}
\label{task-collaborative-decoder}
In previous methods~\cite{moon2023correlation,moon2023query, sun2024tr}, input-agnostic decoder queries are randomly initialized for position or set to zeros for semantics, which neglects spatial semantics and temporal position guidance from the video-text inputs to the decoder. While some efforts~\cite{jang2023knowing} have sought to enhance performance by integrating video features into the decoder queries, these methods suffer from overly complex module designs that reduce practicality and scalability. Actually, the semantics salience predicted along the temporal dimension in HD, can be a prior preference guidance for the MR task. 
The precise temporal boundaries and semantic foregrounds in the MR task could supplement accurate constraints besides subjective ratings in HD.  
To effectively utilize the task-wise correlation in motion-semantics dimensions, we propose the Mutual Task-Collaborated Decoder.

\noindent\textbf{Highlight guides the preference for MR.}
Here we introduce the salience prediction results by HD as prior knowledge to enhance spatial semantics content and temporal position information in the decoder queries. 
Specifically, we select the largest top-{$K$} salience scores from encoder outputs {$X'= X \setminus \{x_s\} = \{x_1, x_2, ..., x_L\}$} as the semantics content queries {$Q_c$}, which can be computed as:
\begin{equation}
    Q_c = \{x_k\in X'|k \in \operatorname{top-K}(S(x_k))\}
    \label{Qc}
\end{equation}
where {$K$} is the number of decoder queries, and the salience score {$S(\cdot)$} is formulated as:
\begin{equation}
   S(x_i) = \frac{\omega_s^T x_s \cdot \omega_v^T x_i}{\sqrt{d}}
  \label{Salience}
\end{equation}
where {$\omega_s$} and { $\omega_v$} are the learnable weights. 
{$x_s \in X$} is the result of the interaction between salience token {$v_s$} and query-based video representations {$\hat{F_v}$} to determine the alignment between video-text pairs, classifying them as matched (positive) or unmatched (negative) and directly contributing to the computation of salience scores.

Furthermore, considering the inherent relationship between spatial semantics content and temporal position information, they consistently occur in proximity within a video.
We transform the {$Q_c$} of Eq.~(\ref{Qc}) into reference position {$R_k$} through an auxiliary span layer, which is an MLP, structurally same as the moment prediction head used in MR but with non-shared parameters, computed as:
\begin{equation}
    R_k = \mbox{MLP}_{\textit{Span}}\left( q_k \right), \quad q_k \in Q_c 
    \label{Rk}
\end{equation}
Then the temporal position queries $Q_{p}$ with $d$-dim is computed as:
\begin{equation}
    \Phi(R_k) = \left( \sin \left(\frac{2\pi R_k}{10000^{2i/\frac{d}{2}}}\right) \oplus \cos \left(\frac{2\pi R_k}{10000^{(2i+1)/\frac{d}{2}}}\right) \right)
    \label{Qp}
\end{equation}
where {$i\in\{0, ..., \frac{d-1}{2}\}$ }. 
Then we combine the $Q_{p}$ and $Q_{c}$ to form the queries input to the Transformer Decoder. These queries with prior knowledge enhance the prediction head's ability to accurately retrieve relevant video moments.

\noindent\textbf{Moments refine the distinction for HD.}
The MR loss { $\mathcal{L}_{MR}$} aims to minimize the discrepancy between retrieval predictions {$\hat{m}$} and its corresponding ground-truth moments $m$, which consists of {$L_{1}$} loss for boundary regression, generalized IoU loss {$\mathcal{L}_{gIoU}$}~\cite{rezatofighi2019generalized} for temporal moments covering, and a binary cross entropy loss {$\mathcal{L}_{ce}$} to differentiate foreground and background prediction $f_i$, which is computed as:
\begin{equation}
    \mathcal{L}_{MR} = \lambda^{MR}_{L1}||m - \hat{m}|| + \lambda^{MR}_{gIoU}\mathcal{L}_{gIoU}\left(m, \hat{m}\right) + \lambda^{MR}_{ce}\mathcal{L}_{ce}(f_i, y_i)
    \label{eq:MR_collab}
\end{equation}
where {$\scriptstyle \lambda^{MR}_{L1}$, $ \scriptstyle \lambda^{MR}_{gIoU}$, $\scriptstyle \lambda^{MR}_{ce}$ }are hyperparameters to balance loss.

These accurate temporal boundaries and semantic foreground constraints can be utilized to enhance the discrimination of salience prediction in HD.
Consequently, by leveraging classification and regression constraint losses from MR task, we improve the learning process in HD.
Specifically, for the salience scores {$\{S(x_i)| i \in \{1,...,L\} \}$} computed by Eq.~(\ref{Salience}), we use the {$\mathcal{L}_{ce}$} to further refine the spatial semantics alignment of relevant moments. 
For the reference position {$R_k$} by Eq.~(\ref{Rk}), we use {$L_{1}$} and {$\mathcal{L}_{gIoU}$} to improve the temporal position accuracy of highlights detected, which is computed by: 
\begin{equation}
\begin{aligned}
    \mathcal{L}_{collab}^{HD} = \lambda^{HD}_{L1}&||m - R_k|| + \lambda^{HD}_{gIoU}\mathcal{L}_{gIoU}\left(m, R_k\right) \\ 
    &+\lambda^{HD}_{ce}\mathcal{L}_{ce}(Sigmoid(S(x_i)), y_i)
    \label{eq:hd_collab}
\end{aligned}
\end{equation}
This strategy leverages MR task's temporal precision and semantics boundary awareness to refine salience predictions in HD, without the need for additional complex components.

\begin{figure*}[t]
  \centering
   \includegraphics[width=0.92\linewidth]{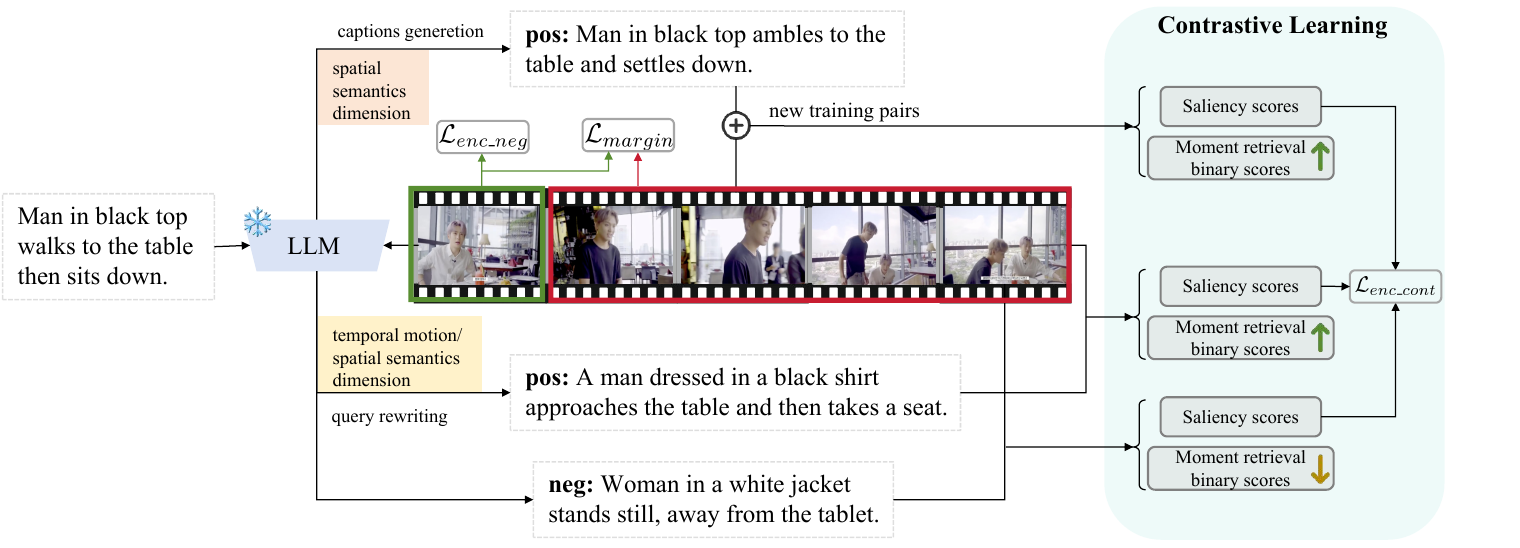}
   \caption{Illustration of Contrastive Metric Learning on Auxiliary Data. We sufficiently utilize the richness of auxiliary data in terms of the motion-semantics dimension to more effectively capture nuanced video content. Video clips in this figure are sampled from the QVHighlight dataset~\cite{lei2021detecting}.} 
   \label{fig:contrastive}
\end{figure*}

\subsection{Contrastive Denoising Training on Auxiliary Data}
While proposed MSDE and MTCD effectively exploit the internal association of motion-semantics dimensions within video, the generalization of the model is constrained by the sparsity dilemma of both dimensions inherent in MR/HD datasets annotations.
For example, QVHighlight~\cite{lei2021detecting} is rich in scenes within a video but many informative scenes are coarsely annotated, lacking detailed spatial semantics information that is critical for understanding visual contexts. 
Conversely, TACoS~\cite{regneri2013grounding} consists of videos in a single scenario and corresponding text descriptions of actions. 
The annotations for these actions are simplistic, lacking detailed temporal motion information that is critical for understanding 
dynamic contexts.

To address this, we design the corpus generation strategy to achieve richness alignment between
video and text at the data level in terms of motion-semantics dimensions.
Besides, we introduce a novel contrastive learning including a contrastive denoising strategy for reliable training and contrastive metric learning to fully utilize the auxiliary data.

\noindent \textbf{Motion-Semantics Corpus Generation}
\label{generate}

\noindent \textit{\textbf{Semantics corpus generation}}
consists of two types: finer captions generation and query rewriting. 
For captions generation, we generate captions for each video clip in the ground truth by pre-trained LLaVA~\cite{liu2024visual} and determine the corresponding moment interval by thresholding a cosine-similarity matrix between CLIP visual and textual embeddings. To ensure meaningful temporal context, captions associated with intervals of a length less than three are filtered out. Subsequently, we select the top-2 caption-video pairs, based on the length of the clip interval to form a new set of video-text pairs. 
For query rewriting, utilizing the language component \cite{touvron2023llama} of LLaVA, the \textit{nouns} in the text queries are replaced by their synonyms or antonyms.
These new queries with their corresponding original video moment intervals are respectively set as positive and hard-negative video-text pairs.

\noindent \textit{\textbf{Motion corpus generation}} is similar to query rewriting. Instead, we replace the \textit{verbs} in text queries with their corresponding original video moment intervals to produce positive and hard-negative video-text pairs.

\noindent \textbf{Contrastive Denoising Learning.} Besides extra motion-semantics knowledge, 
the corpus generation strategy inevitably introduces noise, 
which may disrupt the learning consistency.
To ensure the above robust and effective training from the generated data, we propose a contrastive denoising learning loss $\mathcal{L}_{cont\_dn}$.

We introduce random perturbation to the GT moments {$m = \{[c_i, s_i], i \in \{1,...,G\} \}$} to generate noisy positive and negative queries for the decoder, where {$G$} is the number of {$m$}. 
Our goal is to accurately identify the GT moments from the outputs generated by noisy positive queries and ensure their non-recognition in outputs from noisy negative queries.
We define random { $\lambda_1 \in [0, 1]$} and {$\lambda_2 \in [1,2]$} as the positive and negative noise scale respectively, then add a random noise {$\triangle c$} and {$\triangle s$} to the center and span of GT moments. Make sure that {$|\triangle c| = \frac{\delta_2 \lambda \cdot s}{2}$} and {$|\triangle s| = \frac{\delta_2 \lambda \cdot s}{2}$}, where {$s$} is the span of GT moments, {$\delta_2$} is the hyperparameter to control noise and {$\lambda \in \{\lambda_1,\lambda_2\}$}. The {$\mathcal{L}_{cont\_dn}$} is formulated as:
\begin{equation}
    \mathcal{L}_{cont\_dn} = \lambda^{DN}_{L1}||m - \tilde{m}|| + \lambda^{DN}_{gIoU}\mathcal{L}_{gIoU}\left(m, \tilde{m}\right) + \lambda^{DN}_{ce}\mathcal{L}_{ce} (f_i, y_i)
    \label{eq:dn}
\end{equation}
where $\tilde{m}$ here is the predictions of moments from the decoder, formulated as:
\begin{equation}
    \tilde{m} = \mbox{MTCD}\left(\Phi \left( m_{noise} \right), X\right)
\end{equation}
where {$X=\{x_s, x_1, x_2, \ldots, x_L\}$} is the memory input for the decoder, {$\Phi(\cdot)$} is computed by eq.~(\ref{Qp}) and {$m_{noise}$} is generated by adding the noise {$\triangle c$} and {$\triangle s$} in GT moments {$m$} as follows: 
\begin{equation}
     m_{noise} = \{[c_i + \triangle c_i, s_i + \triangle s_i]\}, i \in \{1,...,G\}
\end{equation}

\begin{table*}[t]
  \centering
  \caption{Performance comparison on QVHighlights test and val splits. We calculate the average mAP score with IoU thresholds ranging from 0.5 to 0.95 in 0.05 intervals.}
  \resizebox{1.0\textwidth}{!}{
  \begin{tabular}{@{}c|ccccc|cc|ccccc|cc@{}}
    \hline
    Split  & \multicolumn{7}{c|}{test} & \multicolumn{7}{c}{val} \\
    \cline{1-15}
    \multirow{3}{*}{Method} & \multicolumn{5}{c|}{MR} & \multicolumn{2}{c|}{HD} & \multicolumn{5}{c|}{MR} & \multicolumn{2}{c}{HD} \\
    \cmidrule(lr){2-6}
    \cmidrule(lr){7-8}
    \cmidrule(lr){9-13}
    \cmidrule(lr){14-15}
    &\multicolumn{3}{c}{mAP} & \multicolumn{2}{c|}{R1} & \multicolumn{2}{c|}{$\geq$ Very Good} & \multicolumn{3}{c}{mAP} & \multicolumn{2}{c|}{R1} & \multicolumn{2}{c}{$\geq$ Very Good} \\
    \cmidrule(lr){2-4}
    \cmidrule(lr){5-6}
    \cmidrule(lr){7-8}
    \cmidrule(lr){9-11}
    \cmidrule(lr){12-13}
    \cmidrule(lr){14-15}
    & @0.5 & @0.75 & Avg & @0.5 & @0.7 & mAP & HIT@1 & @0.5 & @0.75 & Avg  & @0.5 & @0.7 & mAP & HIT@1 \\
    \hline
    BeautyThumb~\cite{song2016click} & - & - & - & - & - & 14.36 & 20.88 & - & - & - & - & - & - & - \\
    DVSE~\cite{liu2015multi}  & - & - & - & - & - & 18.75 & 21.79 & - & - & - & - & - & - & - \\
    MCN~\cite{anne2017localizing} & 24.94 & 8.22 & 10.67 & 11.41 & 2.72 &  - & - & - & - & - & - & - & - & - \\
    CAL~\cite{escorcia2019temporal} & 23.40 & 7.65 & 9.89 & 25.49 & 11.54 & - & - & - & - & - & - & - & - & - \\
    XML~\cite{lei2020tvr} & 44.63 & 31.73 & 32.14 & 41.83 & 30.35 & 34.49 & 55.25 & - & - & - & - & - & - & - \\
    XML+~\cite{lei2020tvr} & 47.89 & 34.67 & 34.90 &  46.69 & 33.46 & 35.38 & 55.06 & - & - & - & - & - & - & - \\
    Moment-DETR~\cite{lei2021detecting} & 54.82 & 29.40 & 30.73 & 52.89 & 33.02 & 35.69 & 55.60 & - & - & 32.20 & 53.94 & 34.84 & 35.65 & 55.55 \\
    UMT~\cite{liu2022umt} & 53.38 & 37.01 & 36.12 & 56.23 & 41.18 & 38.18 & 59.99 & - & - & 38.59 & 60.26 & 44.26 & 39.85 & 64.19 \\
    QD-DETR~\cite{moon2023query} & 62.52 & 39.88 & 39.86 & 62.40 & 44.98 & 38.94 & 62.40 & 62.23 & 41.82 & 41.22 & 62.68 & 46.66 & 39.13 & 63.03 \\
    UniVGT~\cite{lin2023univtg} & 57.6 & 35.59 & 35.47 & 58.86 & 40.86 & 38.20 & 60.96 & - & - & 36.13 & 59.74 & - & 38.83 & 61.81 \\
    EaTR~\cite{jang2023knowing} & - & - & - & - & - & - & - & 61.86 & 41.91 & 41.74 & 61.36 & 45.79 & 37.15 & 58.65 \\
    CG-DETR~\cite{moon2023correlation} & 64.51 & 42.77 & 42.86 & 65.43 & 48.38 & 40.33 & \textbf{66.21} & 64.07 & 42.81 & 42.33 & 64.13 & 48.06 & 39.91 & 64.84 \\
    TR-DETR~\cite{sun2024tr} & 63.98 & 43.73 & 42.62 & 64.66 & \textbf{48.96} & 39.91 & 63.42 & - & - & - & - & - & - & - \\
    \cline{1-15}
    \textbf{MS-DETR (Ours)} & \textbf{66.41} & \textbf{44.91} & \textbf{44.89} & \textbf{64.72} & \underline{48.77} & \textbf{40.45} & \underline{65.95} & \textbf{67.19} & \textbf{46.05} & \textbf{46.00} & \textbf{66.9} & \textbf{51.68} & \textbf{40.57} & \textbf{66.58} \\
    \hline
  \end{tabular}
  }
  \label{tab:qvhighlight}
\end{table*}

\begin{table}[ht]
    \centering
    \caption{Results on MR datasets TACoS test and val splits. Video features are extracted using Slowfast and CLIP.}
    \resizebox{0.45\textwidth}{!}{ 
    \begin{tabular}{c|ccccc}
        \hline
        Split & Method & R@0.3 & R@0.5 & R@0.7 & mIoU \\
        \hline
        \multirow{6}{*}{val} & 2D-TAN~\cite{zhang2020learning} & 40.01 & 27.99 & 12.92 & 27.22 \\
        & VSLNet~\cite{zhang2020span} & 35.54 & 23.54 & 13.15 & 24.99 \\
        & Moment-DETR~\cite{lei2021detecting} & 37.97 & 24.67 & 11.97 & 25.49 \\
        & UniVTG~\cite{lin2023univtg} & 51.44 & 34.97 & 17.35 & 33.60 \\
        & CG-DETR~\cite{moon2023correlation} & 52.23 & 39.61 & 22.23 & 36.48 \\
        & \textbf{MS-DETR (ours)} & \textbf{53.16} & \textbf{39.65} & \textbf{23.42} & \textbf{37.01} \\
        \cline{1-6}
        \multirow{2}{*}{test} & CG-DETR~\cite{moon2023correlation} & 52.23 & 39.61 & 22.23 & 36.48 \\
        & \textbf{MS-DETR (ours)} & \textbf{56.51} & \textbf{43.00} & \textbf{25.37} & \textbf{39.23} \\
        \hline
    \end{tabular}
    }
    \label{tab:tacos}
\end{table}

\noindent \textbf{Contrastive Metric Learning.} The contrastive metric learning consists of three components as shown in Fig.~\ref{fig:contrastive}. We define $X_{neg}$ as the output of the negative video-text pairs from MSDE. 
First, we employ a loss function $\mathcal{L}_{enc\_neg}$ to suppress the salience scores of negative pairs $X_{neg}$ as follows:
\begin{equation}
    \mathcal{L}_{enc\_neg} = -\sum_{x_i \in X_{neg}} \log \left( 1 - Sigmoid\left(S\left(x_i \right)\right) \right)
\end{equation}
Then, the margin loss {$\mathcal{L}_{margin}$} is computed between two pairs of positive and negative clips. The first pair consists of one high score clip (with index {$x_{h}$}) and one low score clip ({$x_{l}$}) within the ground-truth moments. The second pair is a clip within ({$x_{in}$}) and a clip outside ({$x_{out}$}) the ground-truth moments, which is formulated as:
\begin{equation}
\begin{aligned}
    \mathcal{L}_{margin} & = \operatorname{max}\left(0, \delta + S(x_{l}) - S(x_{h})\right) \\
    &+ \operatorname{max}\left(0, \delta + S(x_{out}) - S(x_{in})\right)
\end{aligned}
\end{equation}
where $\delta$ is the margin set as 0.2 by default.
Lastly, a contrastive learning loss is designed to amplify the salience scores of positive queries while concurrently suppressing the salience scores of negative queries for the same video, which is formulated as:
\begin{equation}
    \mathcal{L}_{enc\_cont} = - \frac{1}{N} \sum_{n=1}^N{\log{\left( \frac{\sum_{x_i \in X_{pos}^n} \exp(S\left( x_i \right) / \xi)}{\sum_{x \in {X_{pos}^n \cup X_{neg}^n}} \exp(S\left( x_i \right) / \xi)} \right)}}
    \label{eq:enc_cont}
\end{equation}
where {$\xi$} is a temperature parameter and {$N$} denotes the maximum rank value. We perform $N$ iterations of the data within a batch, and in each iteration {$n \in \{1,2,\ldots,N\}$}, the samples with higher salience scores than {$n$} are used as $X^n_{pos}$ and the lower are set as {$X^n_{neg}$}. Note that the output of negative video-text pairs, \textit{i.e.}, {$X_{neg}$} are also be included in {$\{X^n_{neg}, n \in \{1,...,N\}\}$}. 

\subsection{Training Loss}
The total loss of our model used for training is as follows:
\begin{equation}
\begin{aligned}
   \mathcal{L}_{total} = &\mathcal{L}^{HD}_{collab} + \mathcal{L}_{MR} + \lambda_1\mathcal{L}_{cont\_dn} \\
   &+ \lambda_2(\mathcal{L}_{enc\_neg} + \mathcal{L}_{margin} + \mathcal{L}_{enc\_cont})
   \label{eq:total_loss}
\end{aligned}
\end{equation}
where $\lambda_1$ and $\lambda_2$ control the weights of last three losses.

\begin{table}[t]
    \centering
    \caption{Results on MR datasets Charades-STA test split. Video features are extracted using Slowfast and CLIP.}
    \resizebox{0.4\textwidth}{!}{
    \begin{tabular}{ccccc}
        \hline
        Method & R@0.3 & R@0.5 & R@0.7 & mIoU \\
        \hline
        2D-TAN~\cite{zhang2020learning} & 58.76 & 46.02 & 27.50 & 41.25 \\
        VSLNet~\cite{zhang2020span} & 60.30 & 42.69 & 24.14 & 41.58 \\
        Moment-DETR~\cite{lei2021detecting} & 65.83 & 52.07 & 30.59 & 45.54 \\
        QD-DETR~\cite{moon2023query} & - & 57.31 & 32.55 & - \\
        LLaViLo~\cite{ma2023llavilo} & - & 55.72 & 33.43 & - \\
        UniVTG~\cite{lin2023univtg} & 70.81 & 58.01 & 35.65 & 50.10 \\
        CG-DETR~\cite{moon2023correlation} & 70.43 & 58.44 & 36.34 & 50.13 \\
        \textbf{MS-DETR (ours)} & \textbf{71.34} & \textbf{59.62} & \textbf{36.48} & \textbf{50.59} \\
        \hline
    \end{tabular}
    }
    \label{tab:charades}
\end{table}

\section{Experiment}

\begin{table*}[t]
  \centering
  \caption{Highlight detection results on TVsum. $\dagger$ denotes the methods that utilize the audio modality.}
  \resizebox{0.8\textwidth}{!}{
  \begin{tabular}{@{}c|cccccccccc|c@{}}
    \hline
    Method & VT & VU & GA & MS & PK & PR & FM & BK & BT & DS & Avg \\
    \hline
    sLSTM~\cite{zhang2016video} & 41.1 & 46.2 & 46.3 & 47.7 & 44.8 & 46.1 & 45.2 & 40.6 & 47.1 & 45.5 & 45.1 \\
    SG~\cite{mahasseni2017unsupervised} & 42.3 & 47.2 & 47.5 & 48.9 & 45.6 & 47.3 & 46.4 & 41.7 & 48.3 & 46.6 & 46.2 \\
    LIM-S~\cite{xiong2019less} & 55.9 & 42.9 & 61.2 & 54.0 & 60.3 & 47.5 & 43.2 & 66.3 & 69.1 & 62.6 & 56.3 \\
    Trailer~\cite{wang2020learning} & 61.3 & 54.6 & 65.7 & 60.8 & 59.1 & 70.1 & 58.2 & 64.7 & 65.6 & 68.1 & 62.8 \\
    SL-Module~\cite{xu2021cross} & 86.5 & 68.7 & 74.9 & 86.2 & 79.0 & 63.2 & 58.9 & 72.6 & 78.9 & 64.0 & 73.3 \\
    QD-DETR~\cite{moon2023query} & 88.2 & 87.4 & 85.6 & 85.0 & 85.8 & 86.9 & 76.4 & 91.3 & 89.2 & 73.7 & 85.0 \\
    UniVGT~\cite{lin2023univtg} & 83.9 & 85.1 & 89.0 & 80.1 & 84.6 & 81.4 & 70.9 & 91.7 & 73.5 & 69.3 & 81.0 \\
    CG-DETR~\cite{moon2023correlation} & 86.9 & 88.8 & 94.8 & 87.7 & 86.7 & \textbf{89.6} & 74.8 & 93.3 & 89.2 & 75.9 & 86.8 \\
    TR-DETR~\cite{sun2024tr} & 89.3 & 93.0 & 94.3 & 85.1 & 88.0 & 88.6 & 80.4 & 91.3 & 89.5 & \textbf{81.6} & 88.1 \\ 
    \cline{1-12}
    MINI-Net~\cite{hong2020mini} $\dagger$ & 80.6 & 68.3 & 78.2 & 81.8 & 78.1 & 65.8 & 57.8 & 75.0 & 80.2 & 65.5 & 73.2 \\
    TGG~\cite{ye2021temporal} $\dagger$ & 85.0 & 71.4 & 81.9 & 78.6 & 80.2 & 75.5 & 71.6 & 77.3 & 78.6 & 68.1 & 76.8 \\
    Joint-VA~\cite{badamdorj2021joint} $\dagger$ & 83.7 & 57.3 & 78.5 & 86.1 & 80.1 & 69.2 & 70.0 & 73.0 & \textbf{97.4} & 67.5 & 76.3 \\
    UMT~\cite{liu2022umt} $\dagger$ & 87.5 & 81.5 & 88.2 & 78.8 & 81.4 & 87.0 & 76.0 & 86.9 & 84.4 & 79.6 & 83.1 \\
    QD-DETR~\cite{moon2023query} $\dagger$ & 87.6 & 91.7 & 90.2 & 88.3 & 84.1 & 88.3 & 78.7 & 91.2 & 87.8 & 77.7 & 86.6 \\
    \cline{1-12}
    \textbf{MS-DETR (Ours)} & \textbf{89.84} & \textbf{93.33} & \textbf{94.79} & \textbf{88.73} & \textbf{88.52} & \underline{89.02} & \textbf{80.5} & \textbf{93.96} & 88.73 & 76.88 & \textbf{88.43} \\
    \hline
  \end{tabular}
  }
  \label{tab:tvsum}
\end{table*}

\begin{table*}[ht]
  \centering
  \caption{Effects of different components on QVHighlights val split.}
  \resizebox{0.65\textwidth}{!}{
  \begin{tabular}{@{}c|cccc|ccccc|cc@{}}
    \hline
    & \multirow{3}{*}{MSDE} & \multirow{3}{*}{MTCD} & \multirow{3}{*}{AD} & \multirow{3}{*}{CDL} & \multicolumn{5}{c|}{MR} & \multicolumn{2}{c}{HD} \\
    \cmidrule(lr){6-10}
    \cmidrule(lr){11-12}
    & & & & & \multicolumn{3}{c}{mAP} & \multicolumn{2}{c|}{R1} & \multicolumn{2}{c}{$\geq$ Very Good} \\
    \cmidrule(lr){6-8}
    \cmidrule(lr){9-10}
    \cmidrule(lr){11-12}
    & & & & & @0.5 & @0.75 & \textbf{Avg} & @0.5 & @0.7 & mAP & HIT@1 \\
    \hline
    (a) & & & & & 61.61 & 41.75 & 40.71 & 62.6 & 47.48 & 39.27 & 61.61 \\
    \cline{1-12}
    (b) & \checkmark & & & & 63.08 &  44.03 & 42.76 & 63.1 & 49.23 & 40.21 & 65.35  \\
    (c) & & \checkmark & & & 62.3 & 43.53 & 42.84 & 62.65 & 47.81 & 39.64 & 63.61  \\
    (d) & & & \checkmark & & 65.21 & 43.36 & 43.27 & 65.35 & 49.42 & 40.42 & 65.55  \\
    (e) & & & & \checkmark & 63.1 & 44.08 & 43.24 & 62.58 & 47.42 & 39.03 & 61.61 \\
    \cline{1-12}
    (f) & \checkmark & \checkmark & & & 63.17 & 44.11 & 43.33 & 64.45 & 48.97 & 39.86 & 63.35 \\
    (g) & & & \checkmark & \checkmark & 65.58 & 44.78 & 44.16 & 65.16 & 48.77 & 40.24 & 64.9  \\
    (h) & \checkmark &  & \checkmark & \checkmark & 66.44 & 45.43 & 44.99 & 65.81 & 50.39 & \textbf{40.72} & 65.87  \\
    (i) & & \checkmark & \checkmark & \checkmark & 65.4 & \textbf{46.3} & 45.51 & 65.23 & 50.39 & 39.86 & 64.26  \\
    \cline{1-12}
    (j) & \checkmark & \checkmark & \checkmark & \checkmark & \textbf{67.19} & \underline{46.05} & \textbf{46.0} & \textbf{66.9} & \textbf{51.68} & \underline{40.57} & \textbf{66.58}  \\
    \hline
  \end{tabular}
    }
  \label{tab:ablation}
\begin{tablenotes} 
\item[*] ~~~~~~~~~~~~~~~~~~~~~~~~~~~~~~~~~~~~AD and CDL refer to Contrastive Denoising Learning and Auxiliary Data respectively.
\end{tablenotes}
\end{table*}

\begin{table}[t]
  \centering
  \caption{The impact of model parameters on the performance.}
  \resizebox{0.48\textwidth}{!}{
  \begin{tabular}{@{}c|ccccc|cc@{}}
    \hline
    \multirow{3}{*}{Encoder Layers} & \multicolumn{5}{c|}{MR} & \multicolumn{2}{c}{HD} \\
    \cmidrule(lr){2-6}
    \cmidrule(lr){7-8}
    & \multicolumn{3}{c|}{mAP} & \multicolumn{2}{c}{R1}  & \multicolumn{2}{c}{$\geq$ Very Good} \\
    \cmidrule(lr){2-4}
    \cmidrule(lr){5-6}
    \cmidrule(lr){7-8}
    & @0.5 & @0.75 & Avg & @0.5 & @0.7 & mAP & HIT@1 \\
    \hline
    Baseline (2CAT \& 2SAT) & 61.61 & 41.75 & 40.71 & 62.6 & 47.48 & 39.27 & 61.61  \\
    \textbf{MS-DETR (2CAT \& 2SAT)} & \underline{62.61} & \underline{42.72} & \underline{42.08}  & \underline{62.77} & \underline{47.61} & \underline{39.93} & \underline{63.1} \\
    Baseline (4CAT \& 2SAT) & 62.09 & 42.96 & 41.13 & 62.71 & 47.87 &  38.69 & 60.65 \\
    \textbf{MS-DETR (4CAT \& 2SAT)} & \textbf{63.08} &  \textbf{44.03} & \textbf{42.76} & \textbf{63.1} & \textbf{49.23} & \textbf{40.21} & \textbf{65.35}  \\
    \hline
  \end{tabular}
  }
  \label{tab:ablation_layers}
\end{table}

\subsection{Dataset and Evaluation Metrics}

To validate the effectiveness, we conduct extensive experiments on four mainstream MR/HD datasets. Specifically, we performed MR on QVHighlights~\cite{lei2021detecting}, Charades-STA~\cite{gao2017tall}, and TACoS~\cite{regneri2013grounding} datasets, and HD on QVHighlights~\cite{lei2021detecting} and TVSum~\cite{song2015tvsum} datasets.
All the video motion-semantics features and text features have been pre-extracted by pretrained Slowfast~\cite{feichtenhofer2019slowfast} and CLIP~\cite{radford2021learning}.
For a fair comparison, we adhere to the provided features and evaluation metrics used in previous works~\cite{lei2021detecting,moon2023query,xu2023mh,moon2023correlation,xiao2023bridging,sun2024tr}. 
For the MR task, we measure performance using Recall@1 with IoU thresholds of 0.5 and 0.7, as well as mean average precision (mAP) at various IoU thresholds. For the HD task, we employed mAP and HIT@1 as evaluation metrics.
Note that mAP in MR serves as the most critical metric in joint HD/MR tasks for comparing models and previous methods also used it as the basis for model evaluation.

\subsection{Experimental Results}

\textbf{Results on joint task datasets QVHighlights.}
Tab.~\ref{tab:qvhighlight} presents a comparison of our results with state-of-the-art (SOTA) methods on the joint tasks of MR/HD.
We report metrics for both test and val dataset splits.
As observed, our proposed method significantly outperforms the previous SOTAs in both MR/HD tasks, improving the mAP metric by 2\% and 3.77\% respectively. 
Specifically, we believe that by learning motion-semantics disentangled representations, we enhance the model's ability to precisely retrieve fine-grained motion and semantics features.

\textbf{Results on MR datasets TACoS and Charades-STA.}
Tab.~\ref{tab:tacos} and Tab.~\ref{tab:charades} show the comparisons on MR datasets TACoS and Charades-STA.
For TACoS, we report metrics for both test and val splits and our method outperforms current SOTAs with notable margins, especially improving R@0.5 by 3.39\% and mIoU by 2.75\% on test split. 
Similarly, for Charades-STA, our method demonstrated superior performance, further validating the robustness and effectiveness of our proposed approach across different datasets.

\begin{figure*}[t]
  \centering
   \includegraphics[scale=0.53]{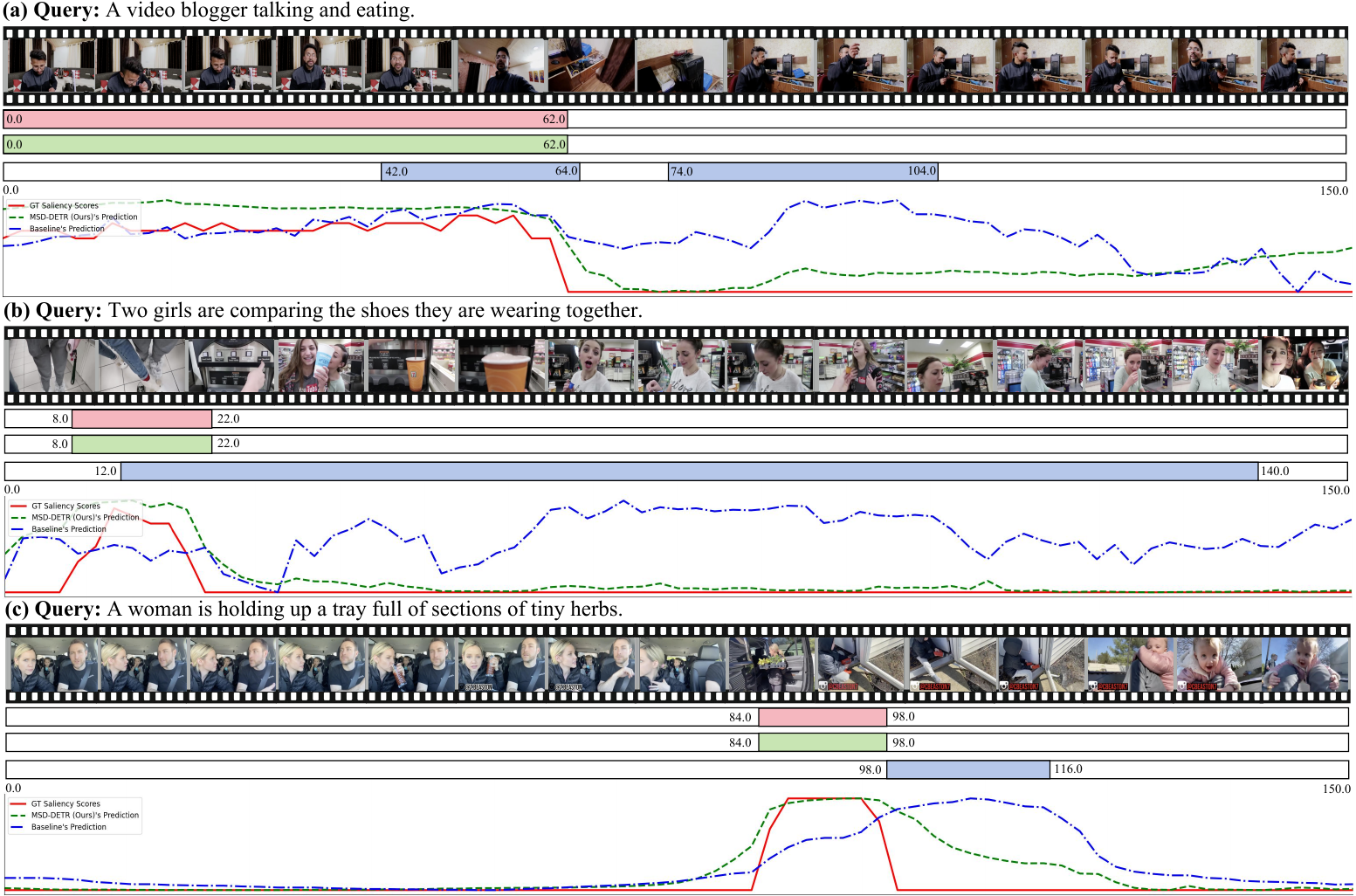}
   \caption{Visualization of prediction comparisons on video clips from the QVHighlight dataset~\cite{lei2021detecting}. Red, green, and blue indicate the qualitative results for ground truth, our MS-DETR, and baseline, respectively. (Best viewed zoomed in on screen)}
   \label{fig:qualitative_result}
\end{figure*}

\textbf{Results on HD dataset TVsum.}
Results for HD benchmark are reported in Tab.~\ref{tab:tvsum}.
Contrary to the MR derived from decoder outputs, salience scores for HD are computed using the encoder outputs. 
It is noteworthy that our approach not only outperforms current SOTAs that solely rely on video features but also surpasses those incorporating additional audio features with notable margins.

\subsection{Ablation study}
We examine the contribution of each major component in the proposed method in Tab.~\ref{tab:ablation}. 
We follow the framework design~\cite{moon2023query} as our baseline model in Exp. (a).
Exp. (b) to (e) show the effectiveness of each component compared to the baseline (a).
(b) ensures the incorporation of motion-semantics contexts into each video clip representation,
(c) fully exploits the potential synergies in temporal motion and spatial semantics dimensions between MR and HD tasks,
and (d) effectively mitigates the sparsity annotation dilemma of both motion-semantics dimensions.
Here solely conducting CDL causes slight degradation on HIT\@1 of HD and R1 in MR task, since CDL is designed not to function independently but to optimally utilize the motion-semantics disentangled representation for auxiliary data. 
Comparing Exp. (b)-(e) with (f)-(i), we can observe that the components complement each other effectively, further validating the robustness and effectiveness of each major component in our proposed approach.

In addition, to verify that the improvement of MSDE is not due to the increase in parameters, we conduct the comparison in Tab.~\ref{tab:ablation_layers}.
We expand the number of cross-attention transformer layers (CAT) and self-attention transformer layers (SAT) of baseline to 4 and 2, and reduce the parameters of MSDE to 2 CAT and 2 SAT, respectively for fair comparison. 
The results indicate that performance gains are due to the MSDE's ability to learn more Fine-Grained Motion-Semantics Representations rather than merely increasing parameters.

\subsection{Qualitative Results}
In Fig.~\ref{fig:qualitative_result}, we visualize the qualitative results of our MS-DETR on the QVHighlights datasets. Compared with baseline~\cite{moon2023query}, MS-DETR shows more reasonable and accurate results in terms of retrieved accuracy and highlight score distribution. In Fig.~\ref{fig:qualitative_result}(a) and (b), our model exhibits an enhanced capability to discern fine-grained temporal motion features, \textit{i.e.}, \"eating\" and \"comparing the shoes\", which substantially improves the precision of moment retrieval.
About Fig.~\ref{fig:qualitative_result}(c), the baseline model struggles to process detailed semantics information, such as \"tiny herbs\", resulting in less precise retrieval outcomes.


\section{Conclusion}

To effectively utilize the correlation between rich motion-semantics information in temporal-spatial embeddings within videos, we propose a unified MR/HD framework in this paper named MS-DETR, which significantly advances the learning of disentangled motion-semantics representations. 
First, we design a Motion-Semantics Disentangled Encoder (MSDE) to explicitly distinguish the temporal motion and spatial semantics dimensions within videos and interact with more refined information in both respective dimensions.
Subsequently, we propose a Mutual Task-Collaborated Decoder (MTCD) to enhance the utilization of motion-semantics features through mutual synergistic task learning.
Last, a novel motion-semantics auxiliary data contrastive learning strategy is proposed to address data sparsity while ensuring robust model training through contrastive denoising.
Extensive experiments on MR/HD benchmarks demonstrate that our method outperforms existing SOTAs.

\textbf{Limitation.} While our proposed MS-DETR effectively leverages rich motion-semantics information within video and outperforms existing SOTAs, a current limitation lies in the incomplete exploitation of multimodal data. Specifically, it does not extensively incorporate other significant modalities like audio, which could enrich context and improve analysis accuracy. This may limit effectiveness in scenarios where audio or other modalities are crucial. Future work will address this by integrating additional modalities to enhance the robustness and applicability of our framework.

\bibliographystyle{ACM-Reference-Format}
\bibliography{main}


\begin{thebibliography}{65}


\ifx \showCODEN    \undefined \def \showCODEN     #1{\unskip}     \fi
\ifx \showDOI      \undefined \def \showDOI       #1{#1}\fi
\ifx \showISBNx    \undefined \def \showISBNx     #1{\unskip}     \fi
\ifx \showISBNxiii \undefined \def \showISBNxiii  #1{\unskip}     \fi
\ifx \showISSN     \undefined \def \showISSN      #1{\unskip}     \fi
\ifx \showLCCN     \undefined \def \showLCCN      #1{\unskip}     \fi
\ifx \shownote     \undefined \def \shownote      #1{#1}          \fi
\ifx \showarticletitle \undefined \def \showarticletitle #1{#1}   \fi
\ifx \showURL      \undefined \def \showURL       {\relax}        \fi
\providecommand\bibfield[2]{#2}
\providecommand\bibinfo[2]{#2}
\providecommand\natexlab[1]{#1}
\providecommand\showeprint[2][]{arXiv:#2}

\bibitem[Anne~Hendricks et~al\mbox{.}(2017)]%
        {anne2017localizing}
\bibfield{author}{\bibinfo{person}{Lisa Anne~Hendricks}, \bibinfo{person}{Oliver Wang}, \bibinfo{person}{Eli Shechtman}, \bibinfo{person}{Josef Sivic}, \bibinfo{person}{Trevor Darrell}, {and} \bibinfo{person}{Bryan Russell}.} \bibinfo{year}{2017}\natexlab{}.
\newblock \showarticletitle{Localizing moments in video with natural language}. In \bibinfo{booktitle}{\emph{Proceedings of the IEEE international conference on computer vision}}. \bibinfo{pages}{5803--5812}.
\newblock


\bibitem[Badamdorj et~al\mbox{.}(2021)]%
        {badamdorj2021joint}
\bibfield{author}{\bibinfo{person}{Taivanbat Badamdorj}, \bibinfo{person}{Mrigank Rochan}, \bibinfo{person}{Yang Wang}, {and} \bibinfo{person}{Li Cheng}.} \bibinfo{year}{2021}\natexlab{}.
\newblock \showarticletitle{Joint visual and audio learning for video highlight detection}. In \bibinfo{booktitle}{\emph{Proceedings of the IEEE/CVF International Conference on Computer Vision}}. \bibinfo{pages}{8127--8137}.
\newblock


\bibitem[Badamdorj et~al\mbox{.}(2022)]%
        {badamdorj2022contrastive}
\bibfield{author}{\bibinfo{person}{Taivanbat Badamdorj}, \bibinfo{person}{Mrigank Rochan}, \bibinfo{person}{Yang Wang}, {and} \bibinfo{person}{Li Cheng}.} \bibinfo{year}{2022}\natexlab{}.
\newblock \showarticletitle{Contrastive learning for unsupervised video highlight detection}. In \bibinfo{booktitle}{\emph{Proceedings of the IEEE/CVF Conference on Computer Vision and Pattern Recognition}}. \bibinfo{pages}{14042--14052}.
\newblock


\bibitem[Cai et~al\mbox{.}(2018)]%
        {cai2018weakly}
\bibfield{author}{\bibinfo{person}{Sijia Cai}, \bibinfo{person}{Wangmeng Zuo}, \bibinfo{person}{Larry~S Davis}, {and} \bibinfo{person}{Lei Zhang}.} \bibinfo{year}{2018}\natexlab{}.
\newblock \showarticletitle{Weakly-supervised video summarization using variational encoder-decoder and web prior}. In \bibinfo{booktitle}{\emph{Proceedings of the European conference on computer vision (ECCV)}}. \bibinfo{pages}{184--200}.
\newblock


\bibitem[Carion et~al\mbox{.}(2020)]%
        {carion2020end}
\bibfield{author}{\bibinfo{person}{Nicolas Carion}, \bibinfo{person}{Francisco Massa}, \bibinfo{person}{Gabriel Synnaeve}, \bibinfo{person}{Nicolas Usunier}, \bibinfo{person}{Alexander Kirillov}, {and} \bibinfo{person}{Sergey Zagoruyko}.} \bibinfo{year}{2020}\natexlab{}.
\newblock \showarticletitle{End-to-end object detection with transformers}. In \bibinfo{booktitle}{\emph{European conference on computer vision}}. Springer, \bibinfo{pages}{213--229}.
\newblock


\bibitem[Chen et~al\mbox{.}(2018)]%
        {chen2018temporally}
\bibfield{author}{\bibinfo{person}{Jingyuan Chen}, \bibinfo{person}{Xinpeng Chen}, \bibinfo{person}{Lin Ma}, \bibinfo{person}{Zequn Jie}, {and} \bibinfo{person}{Tat-Seng Chua}.} \bibinfo{year}{2018}\natexlab{}.
\newblock \showarticletitle{Temporally grounding natural sentence in video}. In \bibinfo{booktitle}{\emph{Proceedings of the 2018 conference on empirical methods in natural language processing}}. \bibinfo{pages}{162--171}.
\newblock


\bibitem[Escorcia et~al\mbox{.}(2019)]%
        {escorcia2019temporal}
\bibfield{author}{\bibinfo{person}{Victor Escorcia}, \bibinfo{person}{Mattia Soldan}, \bibinfo{person}{Josef Sivic}, \bibinfo{person}{Bernard Ghanem}, {and} \bibinfo{person}{Bryan Russell}.} \bibinfo{year}{2019}\natexlab{}.
\newblock \showarticletitle{Temporal localization of moments in video collections with natural language}.
\newblock  (\bibinfo{year}{2019}).
\newblock


\bibitem[Feichtenhofer et~al\mbox{.}(2019)]%
        {feichtenhofer2019slowfast}
\bibfield{author}{\bibinfo{person}{Christoph Feichtenhofer}, \bibinfo{person}{Haoqi Fan}, \bibinfo{person}{Jitendra Malik}, {and} \bibinfo{person}{Kaiming He}.} \bibinfo{year}{2019}\natexlab{}.
\newblock \showarticletitle{Slowfast networks for video recognition}. In \bibinfo{booktitle}{\emph{Proceedings of the IEEE/CVF international conference on computer vision}}. \bibinfo{pages}{6202--6211}.
\newblock


\bibitem[Gao et~al\mbox{.}(2017)]%
        {gao2017tall}
\bibfield{author}{\bibinfo{person}{Jiyang Gao}, \bibinfo{person}{Chen Sun}, \bibinfo{person}{Zhenheng Yang}, {and} \bibinfo{person}{Ram Nevatia}.} \bibinfo{year}{2017}\natexlab{}.
\newblock \showarticletitle{Tall: Temporal activity localization via language query}. In \bibinfo{booktitle}{\emph{Proceedings of the IEEE international conference on computer vision}}. \bibinfo{pages}{5267--5275}.
\newblock


\bibitem[Ge et~al\mbox{.}(2019)]%
        {ge2019mac}
\bibfield{author}{\bibinfo{person}{Runzhou Ge}, \bibinfo{person}{Jiyang Gao}, \bibinfo{person}{Kan Chen}, {and} \bibinfo{person}{Ram Nevatia}.} \bibinfo{year}{2019}\natexlab{}.
\newblock \showarticletitle{Mac: Mining activity concepts for language-based temporal localization}. In \bibinfo{booktitle}{\emph{2019 IEEE winter conference on applications of computer vision (WACV)}}. IEEE, \bibinfo{pages}{245--253}.
\newblock


\bibitem[Ghosh et~al\mbox{.}(2019)]%
        {ghosh2019excl}
\bibfield{author}{\bibinfo{person}{Soham Ghosh}, \bibinfo{person}{Anuva Agarwal}, \bibinfo{person}{Zarana Parekh}, {and} \bibinfo{person}{Alexander Hauptmann}.} \bibinfo{year}{2019}\natexlab{}.
\newblock \showarticletitle{Excl: Extractive clip localization using natural language descriptions}.
\newblock \bibinfo{journal}{\emph{arXiv preprint arXiv:1904.02755}} (\bibinfo{year}{2019}).
\newblock


\bibitem[Gygli et~al\mbox{.}(2016)]%
        {gygli2016video2gif}
\bibfield{author}{\bibinfo{person}{Michael Gygli}, \bibinfo{person}{Yale Song}, {and} \bibinfo{person}{Liangliang Cao}.} \bibinfo{year}{2016}\natexlab{}.
\newblock \showarticletitle{Video2gif: Automatic generation of animated gifs from video}. In \bibinfo{booktitle}{\emph{Proceedings of the IEEE conference on computer vision and pattern recognition}}. \bibinfo{pages}{1001--1009}.
\newblock


\bibitem[Hong et~al\mbox{.}(2020)]%
        {hong2020mini}
\bibfield{author}{\bibinfo{person}{Fa-Ting Hong}, \bibinfo{person}{Xuanteng Huang}, \bibinfo{person}{Wei-Hong Li}, {and} \bibinfo{person}{Wei-Shi Zheng}.} \bibinfo{year}{2020}\natexlab{}.
\newblock \showarticletitle{Mini-net: Multiple instance ranking network for video highlight detection}. In \bibinfo{booktitle}{\emph{Computer Vision--ECCV 2020: 16th European Conference, Glasgow, UK, August 23--28, 2020, Proceedings, Part XIII 16}}. Springer, \bibinfo{pages}{345--360}.
\newblock


\bibitem[Jang et~al\mbox{.}(2023)]%
        {jang2023knowing}
\bibfield{author}{\bibinfo{person}{Jinhyun Jang}, \bibinfo{person}{Jungin Park}, \bibinfo{person}{Jin Kim}, \bibinfo{person}{Hyeongjun Kwon}, {and} \bibinfo{person}{Kwanghoon Sohn}.} \bibinfo{year}{2023}\natexlab{}.
\newblock \showarticletitle{Knowing where to focus: Event-aware transformer for video grounding}. In \bibinfo{booktitle}{\emph{Proceedings of the IEEE/CVF International Conference on Computer Vision}}. \bibinfo{pages}{13846--13856}.
\newblock


\bibitem[Jia et~al\mbox{.}(2021)]%
        {jia2021scaling}
\bibfield{author}{\bibinfo{person}{Chao Jia}, \bibinfo{person}{Yinfei Yang}, \bibinfo{person}{Ye Xia}, \bibinfo{person}{Yi-Ting Chen}, \bibinfo{person}{Zarana Parekh}, \bibinfo{person}{Hieu Pham}, \bibinfo{person}{Quoc Le}, \bibinfo{person}{Yun-Hsuan Sung}, \bibinfo{person}{Zhen Li}, {and} \bibinfo{person}{Tom Duerig}.} \bibinfo{year}{2021}\natexlab{}.
\newblock \showarticletitle{Scaling up visual and vision-language representation learning with noisy text supervision}. In \bibinfo{booktitle}{\emph{International conference on machine learning}}. PMLR, \bibinfo{pages}{4904--4916}.
\newblock


\bibitem[Khosla et~al\mbox{.}(2013)]%
        {khosla2013large}
\bibfield{author}{\bibinfo{person}{Aditya Khosla}, \bibinfo{person}{Raffay Hamid}, \bibinfo{person}{Chih-Jen Lin}, {and} \bibinfo{person}{Neel Sundaresan}.} \bibinfo{year}{2013}\natexlab{}.
\newblock \showarticletitle{Large-scale video summarization using web-image priors}. In \bibinfo{booktitle}{\emph{Proceedings of the IEEE conference on computer vision and pattern recognition}}. \bibinfo{pages}{2698--2705}.
\newblock


\bibitem[Kudi and Namboodiri(2017)]%
        {kudi2017words}
\bibfield{author}{\bibinfo{person}{Sukanya Kudi} {and} \bibinfo{person}{Anoop~M Namboodiri}.} \bibinfo{year}{2017}\natexlab{}.
\newblock \showarticletitle{Words speak for actions: Using text to find video highlights}. In \bibinfo{booktitle}{\emph{2017 4th IAPR Asian Conference on Pattern Recognition (ACPR)}}. IEEE, \bibinfo{pages}{322--327}.
\newblock


\bibitem[Lei et~al\mbox{.}(2021)]%
        {lei2021detecting}
\bibfield{author}{\bibinfo{person}{Jie Lei}, \bibinfo{person}{Tamara~L Berg}, {and} \bibinfo{person}{Mohit Bansal}.} \bibinfo{year}{2021}\natexlab{}.
\newblock \showarticletitle{Detecting moments and highlights in videos via natural language queries}.
\newblock \bibinfo{journal}{\emph{Advances in Neural Information Processing Systems}}  \bibinfo{volume}{34} (\bibinfo{year}{2021}), \bibinfo{pages}{11846--11858}.
\newblock


\bibitem[Lei et~al\mbox{.}(2020)]%
        {lei2020tvr}
\bibfield{author}{\bibinfo{person}{Jie Lei}, \bibinfo{person}{Licheng Yu}, \bibinfo{person}{Tamara~L Berg}, {and} \bibinfo{person}{Mohit Bansal}.} \bibinfo{year}{2020}\natexlab{}.
\newblock \showarticletitle{Tvr: A large-scale dataset for video-subtitle moment retrieval}. In \bibinfo{booktitle}{\emph{Computer Vision--ECCV 2020: 16th European Conference, Glasgow, UK, August 23--28, 2020, Proceedings, Part XXI 16}}. Springer, \bibinfo{pages}{447--463}.
\newblock


\bibitem[Li et~al\mbox{.}(2022)]%
        {li2022blip}
\bibfield{author}{\bibinfo{person}{Junnan Li}, \bibinfo{person}{Dongxu Li}, \bibinfo{person}{Caiming Xiong}, {and} \bibinfo{person}{Steven Hoi}.} \bibinfo{year}{2022}\natexlab{}.
\newblock \showarticletitle{Blip: Bootstrapping language-image pre-training for unified vision-language understanding and generation}. In \bibinfo{booktitle}{\emph{International conference on machine learning}}. PMLR, \bibinfo{pages}{12888--12900}.
\newblock


\bibitem[Li et~al\mbox{.}(2025a)]%
        {li2025visuals}
\bibfield{author}{\bibinfo{person}{Mingxiao Li}, \bibinfo{person}{Fang Qu}, \bibinfo{person}{Zhanpeng Chen}, \bibinfo{person}{Na Su}, \bibinfo{person}{Zhizhou Zhong}, \bibinfo{person}{Ziyang Chen}, \bibinfo{person}{Nan Du}, {and} \bibinfo{person}{Xiaolong Li}.} \bibinfo{year}{2025}\natexlab{a}.
\newblock \showarticletitle{From Visuals to Vocabulary: Establishing Equivalence Between Image and Text Token Through Autoregressive Pre-training in MLLMs}.
\newblock \bibinfo{journal}{\emph{arXiv preprint arXiv:2502.09093}} (\bibinfo{year}{2025}).
\newblock


\bibitem[Li et~al\mbox{.}(2025b)]%
        {li2025vista}
\bibfield{author}{\bibinfo{person}{Mingxiao Li}, \bibinfo{person}{Na Su}, \bibinfo{person}{Fang Qu}, \bibinfo{person}{Zhizhou Zhong}, \bibinfo{person}{Ziyang Chen}, \bibinfo{person}{Yuan Li}, \bibinfo{person}{Zhaopeng Tu}, {and} \bibinfo{person}{Xiaolong Li}.} \bibinfo{year}{2025}\natexlab{b}.
\newblock \showarticletitle{VISTA: Enhancing Vision-Text Alignment in MLLMs via Cross-Modal Mutual Information Maximization}.
\newblock \bibinfo{journal}{\emph{arXiv preprint arXiv:2505.10917}} (\bibinfo{year}{2025}).
\newblock


\bibitem[Lin et~al\mbox{.}(2023)]%
        {lin2023univtg}
\bibfield{author}{\bibinfo{person}{Kevin~Qinghong Lin}, \bibinfo{person}{Pengchuan Zhang}, \bibinfo{person}{Joya Chen}, \bibinfo{person}{Shraman Pramanick}, \bibinfo{person}{Difei Gao}, \bibinfo{person}{Alex~Jinpeng Wang}, \bibinfo{person}{Rui Yan}, {and} \bibinfo{person}{Mike~Zheng Shou}.} \bibinfo{year}{2023}\natexlab{}.
\newblock \showarticletitle{Univtg: Towards unified video-language temporal grounding}. In \bibinfo{booktitle}{\emph{Proceedings of the IEEE/CVF International Conference on Computer Vision}}. \bibinfo{pages}{2794--2804}.
\newblock


\bibitem[Liu et~al\mbox{.}(2024)]%
        {liu2024visual}
\bibfield{author}{\bibinfo{person}{Haotian Liu}, \bibinfo{person}{Chunyuan Li}, \bibinfo{person}{Qingyang Wu}, {and} \bibinfo{person}{Yong~Jae Lee}.} \bibinfo{year}{2024}\natexlab{}.
\newblock \showarticletitle{Visual instruction tuning}.
\newblock \bibinfo{journal}{\emph{Advances in neural information processing systems}}  \bibinfo{volume}{36} (\bibinfo{year}{2024}).
\newblock


\bibitem[Liu et~al\mbox{.}(2018)]%
        {liu2018cross}
\bibfield{author}{\bibinfo{person}{Meng Liu}, \bibinfo{person}{Xiang Wang}, \bibinfo{person}{Liqiang Nie}, \bibinfo{person}{Qi Tian}, \bibinfo{person}{Baoquan Chen}, {and} \bibinfo{person}{Tat-Seng Chua}.} \bibinfo{year}{2018}\natexlab{}.
\newblock \showarticletitle{Cross-modal moment localization in videos}. In \bibinfo{booktitle}{\emph{Proceedings of the 26th ACM international conference on Multimedia}}. \bibinfo{pages}{843--851}.
\newblock


\bibitem[Liu et~al\mbox{.}(2015)]%
        {liu2015multi}
\bibfield{author}{\bibinfo{person}{Wu Liu}, \bibinfo{person}{Tao Mei}, \bibinfo{person}{Yongdong Zhang}, \bibinfo{person}{Cherry Che}, {and} \bibinfo{person}{Jiebo Luo}.} \bibinfo{year}{2015}\natexlab{}.
\newblock \showarticletitle{Multi-task deep visual-semantic embedding for video thumbnail selection}. In \bibinfo{booktitle}{\emph{Proceedings of the IEEE conference on computer vision and pattern recognition}}. \bibinfo{pages}{3707--3715}.
\newblock


\bibitem[Liu et~al\mbox{.}(2022)]%
        {liu2022umt}
\bibfield{author}{\bibinfo{person}{Ye Liu}, \bibinfo{person}{Siyuan Li}, \bibinfo{person}{Yang Wu}, \bibinfo{person}{Chang-Wen Chen}, \bibinfo{person}{Ying Shan}, {and} \bibinfo{person}{Xiaohu Qie}.} \bibinfo{year}{2022}\natexlab{}.
\newblock \showarticletitle{Umt: Unified multi-modal transformers for joint video moment retrieval and highlight detection}. In \bibinfo{booktitle}{\emph{Proceedings of the IEEE/CVF Conference on Computer Vision and Pattern Recognition}}. \bibinfo{pages}{3042--3051}.
\newblock


\bibitem[Lu et~al\mbox{.}(2024)]%
        {lu2024coarse}
\bibfield{author}{\bibinfo{person}{Yanzuo Lu}, \bibinfo{person}{Manlin Zhang}, \bibinfo{person}{Andy~J Ma}, \bibinfo{person}{Xiaohua Xie}, {and} \bibinfo{person}{Jianhuang Lai}.} \bibinfo{year}{2024}\natexlab{}.
\newblock \showarticletitle{Coarse-to-fine latent diffusion for pose-guided person image synthesis}. In \bibinfo{booktitle}{\emph{Proceedings of the IEEE/CVF Conference on Computer Vision and Pattern Recognition}}. \bibinfo{pages}{6420--6429}.
\newblock


\bibitem[Ma et~al\mbox{.}(2025)]%
        {ma2025generativeregressionbasedwatch}
\bibfield{author}{\bibinfo{person}{Hongxu Ma}, \bibinfo{person}{Kai Tian}, \bibinfo{person}{Tao Zhang}, \bibinfo{person}{Xuefeng Zhang}, \bibinfo{person}{Han Zhou}, \bibinfo{person}{Chunjie Chen}, \bibinfo{person}{Han Li}, \bibinfo{person}{Jihong Guan}, {and} \bibinfo{person}{Shuigeng Zhou}.} \bibinfo{year}{2025}\natexlab{}.
\newblock \bibinfo{title}{Generative Regression Based Watch Time Prediction for Short-Video Recommendation}.
\newblock
\newblock
\showeprint[arxiv]{2412.20211}~[cs.LG]


\bibitem[Ma et~al\mbox{.}(2023)]%
        {ma2023llavilo}
\bibfield{author}{\bibinfo{person}{Kaijing Ma}, \bibinfo{person}{Xianghao Zang}, \bibinfo{person}{Zerun Feng}, \bibinfo{person}{Han Fang}, \bibinfo{person}{Chao Ban}, \bibinfo{person}{Yuhan Wei}, \bibinfo{person}{Zhongjiang He}, \bibinfo{person}{Yongxiang Li}, {and} \bibinfo{person}{Hao Sun}.} \bibinfo{year}{2023}\natexlab{}.
\newblock \showarticletitle{LLaViLo: Boosting Video Moment Retrieval via Adapter-Based Multimodal Modeling}. In \bibinfo{booktitle}{\emph{Proceedings of the IEEE/CVF International Conference on Computer Vision}}. \bibinfo{pages}{2798--2803}.
\newblock


\bibitem[Mahasseni et~al\mbox{.}(2017)]%
        {mahasseni2017unsupervised}
\bibfield{author}{\bibinfo{person}{Behrooz Mahasseni}, \bibinfo{person}{Michael Lam}, {and} \bibinfo{person}{Sinisa Todorovic}.} \bibinfo{year}{2017}\natexlab{}.
\newblock \showarticletitle{Unsupervised video summarization with adversarial lstm networks}. In \bibinfo{booktitle}{\emph{Proceedings of the IEEE conference on Computer Vision and Pattern Recognition}}. \bibinfo{pages}{202--211}.
\newblock


\bibitem[Mi et~al\mbox{.}(2025)]%
        {mi2025data}
\bibfield{author}{\bibinfo{person}{Yuxi Mi}, \bibinfo{person}{Zhizhou Zhong}, \bibinfo{person}{Yuge Huang}, \bibinfo{person}{Qiuyang Yuan}, \bibinfo{person}{Xuan Zhao}, \bibinfo{person}{Jianqing Xu}, \bibinfo{person}{Shouhong Ding}, \bibinfo{person}{Shaoming Wang}, \bibinfo{person}{Rizen Guo}, {and} \bibinfo{person}{Shuigeng Zhou}.} \bibinfo{year}{2025}\natexlab{}.
\newblock \showarticletitle{Data Synthesis with Diverse Styles for Face Recognition via 3DMM-Guided Diffusion}. In \bibinfo{booktitle}{\emph{Proceedings of the Computer Vision and Pattern Recognition Conference}}. \bibinfo{pages}{21203--21214}.
\newblock


\bibitem[Moon et~al\mbox{.}(2023a)]%
        {moon2023correlation}
\bibfield{author}{\bibinfo{person}{WonJun Moon}, \bibinfo{person}{Sangeek Hyun}, \bibinfo{person}{SuBeen Lee}, {and} \bibinfo{person}{Jae-Pil Heo}.} \bibinfo{year}{2023}\natexlab{a}.
\newblock \showarticletitle{Correlation-guided Query-Dependency Calibration in Video Representation Learning for Temporal Grounding}.
\newblock \bibinfo{journal}{\emph{arXiv preprint arXiv:2311.08835}} (\bibinfo{year}{2023}).
\newblock


\bibitem[Moon et~al\mbox{.}(2023b)]%
        {moon2023query}
\bibfield{author}{\bibinfo{person}{WonJun Moon}, \bibinfo{person}{Sangeek Hyun}, \bibinfo{person}{SangUk Park}, \bibinfo{person}{Dongchan Park}, {and} \bibinfo{person}{Jae-Pil Heo}.} \bibinfo{year}{2023}\natexlab{b}.
\newblock \showarticletitle{Query-dependent video representation for moment retrieval and highlight detection}. In \bibinfo{booktitle}{\emph{Proceedings of the IEEE/CVF Conference on Computer Vision and Pattern Recognition}}. \bibinfo{pages}{23023--23033}.
\newblock


\bibitem[Mun et~al\mbox{.}(2020)]%
        {mun2020local}
\bibfield{author}{\bibinfo{person}{Jonghwan Mun}, \bibinfo{person}{Minsu Cho}, {and} \bibinfo{person}{Bohyung Han}.} \bibinfo{year}{2020}\natexlab{}.
\newblock \showarticletitle{Local-global video-text interactions for temporal grounding}. In \bibinfo{booktitle}{\emph{Proceedings of the IEEE/CVF Conference on Computer Vision and Pattern Recognition}}. \bibinfo{pages}{10810--10819}.
\newblock


\bibitem[Panda et~al\mbox{.}(2017)]%
        {panda2017weakly}
\bibfield{author}{\bibinfo{person}{Rameswar Panda}, \bibinfo{person}{Abir Das}, \bibinfo{person}{Ziyan Wu}, \bibinfo{person}{Jan Ernst}, {and} \bibinfo{person}{Amit~K Roy-Chowdhury}.} \bibinfo{year}{2017}\natexlab{}.
\newblock \showarticletitle{Weakly supervised summarization of web videos}. In \bibinfo{booktitle}{\emph{Proceedings of the IEEE international conference on computer vision}}. \bibinfo{pages}{3657--3666}.
\newblock


\bibitem[Radford et~al\mbox{.}(2021)]%
        {radford2021learning}
\bibfield{author}{\bibinfo{person}{Alec Radford}, \bibinfo{person}{Jong~Wook Kim}, \bibinfo{person}{Chris Hallacy}, \bibinfo{person}{Aditya Ramesh}, \bibinfo{person}{Gabriel Goh}, \bibinfo{person}{Sandhini Agarwal}, \bibinfo{person}{Girish Sastry}, \bibinfo{person}{Amanda Askell}, \bibinfo{person}{Pamela Mishkin}, \bibinfo{person}{Jack Clark}, {et~al\mbox{.}}} \bibinfo{year}{2021}\natexlab{}.
\newblock \showarticletitle{Learning transferable visual models from natural language supervision}. In \bibinfo{booktitle}{\emph{International conference on machine learning}}. PMLR, \bibinfo{pages}{8748--8763}.
\newblock


\bibitem[Regneri et~al\mbox{.}(2013)]%
        {regneri2013grounding}
\bibfield{author}{\bibinfo{person}{Michaela Regneri}, \bibinfo{person}{Marcus Rohrbach}, \bibinfo{person}{Dominikus Wetzel}, \bibinfo{person}{Stefan Thater}, \bibinfo{person}{Bernt Schiele}, {and} \bibinfo{person}{Manfred Pinkal}.} \bibinfo{year}{2013}\natexlab{}.
\newblock \showarticletitle{Grounding action descriptions in videos}.
\newblock \bibinfo{journal}{\emph{Transactions of the Association for Computational Linguistics}}  \bibinfo{volume}{1} (\bibinfo{year}{2013}), \bibinfo{pages}{25--36}.
\newblock


\bibitem[Rezatofighi et~al\mbox{.}(2019)]%
        {rezatofighi2019generalized}
\bibfield{author}{\bibinfo{person}{Hamid Rezatofighi}, \bibinfo{person}{Nathan Tsoi}, \bibinfo{person}{JunYoung Gwak}, \bibinfo{person}{Amir Sadeghian}, \bibinfo{person}{Ian Reid}, {and} \bibinfo{person}{Silvio Savarese}.} \bibinfo{year}{2019}\natexlab{}.
\newblock \showarticletitle{Generalized intersection over union: A metric and a loss for bounding box regression}. In \bibinfo{booktitle}{\emph{Proceedings of the IEEE/CVF conference on computer vision and pattern recognition}}. \bibinfo{pages}{658--666}.
\newblock


\bibitem[Rochan et~al\mbox{.}(2018)]%
        {rochan2018video}
\bibfield{author}{\bibinfo{person}{Mrigank Rochan}, \bibinfo{person}{Linwei Ye}, {and} \bibinfo{person}{Yang Wang}.} \bibinfo{year}{2018}\natexlab{}.
\newblock \showarticletitle{Video summarization using fully convolutional sequence networks}. In \bibinfo{booktitle}{\emph{Proceedings of the European conference on computer vision (ECCV)}}. \bibinfo{pages}{347--363}.
\newblock


\bibitem[Song et~al\mbox{.}(2016)]%
        {song2016click}
\bibfield{author}{\bibinfo{person}{Yale Song}, \bibinfo{person}{Miriam Redi}, \bibinfo{person}{Jordi Vallmitjana}, {and} \bibinfo{person}{Alejandro Jaimes}.} \bibinfo{year}{2016}\natexlab{}.
\newblock \showarticletitle{To click or not to click: Automatic selection of beautiful thumbnails from videos}. In \bibinfo{booktitle}{\emph{Proceedings of the 25th ACM international on conference on information and knowledge management}}. \bibinfo{pages}{659--668}.
\newblock


\bibitem[Song et~al\mbox{.}(2015)]%
        {song2015tvsum}
\bibfield{author}{\bibinfo{person}{Yale Song}, \bibinfo{person}{Jordi Vallmitjana}, \bibinfo{person}{Amanda Stent}, {and} \bibinfo{person}{Alejandro Jaimes}.} \bibinfo{year}{2015}\natexlab{}.
\newblock \showarticletitle{Tvsum: Summarizing web videos using titles}. In \bibinfo{booktitle}{\emph{Proceedings of the IEEE conference on computer vision and pattern recognition}}. \bibinfo{pages}{5179--5187}.
\newblock


\bibitem[Sun et~al\mbox{.}(2024)]%
        {sun2024tr}
\bibfield{author}{\bibinfo{person}{Hao Sun}, \bibinfo{person}{Mingyao Zhou}, \bibinfo{person}{Wenjing Chen}, {and} \bibinfo{person}{Wei Xie}.} \bibinfo{year}{2024}\natexlab{}.
\newblock \showarticletitle{TR-DETR: Task-Reciprocal Transformer for Joint Moment Retrieval and Highlight Detection}.
\newblock \bibinfo{journal}{\emph{arXiv preprint arXiv:2401.02309}} (\bibinfo{year}{2024}).
\newblock


\bibitem[Sun et~al\mbox{.}(2014)]%
        {sun2014ranking}
\bibfield{author}{\bibinfo{person}{Min Sun}, \bibinfo{person}{Ali Farhadi}, {and} \bibinfo{person}{Steve Seitz}.} \bibinfo{year}{2014}\natexlab{}.
\newblock \showarticletitle{Ranking domain-specific highlights by analyzing edited videos}. In \bibinfo{booktitle}{\emph{Computer Vision--ECCV 2014: 13th European Conference, Zurich, Switzerland, September 6-12, 2014, Proceedings, Part I 13}}. Springer, \bibinfo{pages}{787--802}.
\newblock


\bibitem[Touvron et~al\mbox{.}(2023)]%
        {touvron2023llama}
\bibfield{author}{\bibinfo{person}{Hugo Touvron}, \bibinfo{person}{Thibaut Lavril}, \bibinfo{person}{Gautier Izacard}, \bibinfo{person}{Xavier Martinet}, \bibinfo{person}{Marie-Anne Lachaux}, \bibinfo{person}{Timoth{\'e}e Lacroix}, \bibinfo{person}{Baptiste Rozi{\`e}re}, \bibinfo{person}{Naman Goyal}, \bibinfo{person}{Eric Hambro}, \bibinfo{person}{Faisal Azhar}, {et~al\mbox{.}}} \bibinfo{year}{2023}\natexlab{}.
\newblock \showarticletitle{Llama: Open and efficient foundation language models}.
\newblock \bibinfo{journal}{\emph{arXiv preprint arXiv:2302.13971}} (\bibinfo{year}{2023}).
\newblock


\bibitem[Wang et~al\mbox{.}(2020)]%
        {wang2020learning}
\bibfield{author}{\bibinfo{person}{Lezi Wang}, \bibinfo{person}{Dong Liu}, \bibinfo{person}{Rohit Puri}, {and} \bibinfo{person}{Dimitris~N Metaxas}.} \bibinfo{year}{2020}\natexlab{}.
\newblock \showarticletitle{Learning trailer moments in full-length movies with co-contrastive attention}. In \bibinfo{booktitle}{\emph{Computer Vision--ECCV 2020: 16th European Conference, Glasgow, UK, August 23--28, 2020, Proceedings, Part XVIII 16}}. Springer, \bibinfo{pages}{300--316}.
\newblock


\bibitem[Xiao et~al\mbox{.}(2023)]%
        {xiao2023bridging}
\bibfield{author}{\bibinfo{person}{Yicheng Xiao}, \bibinfo{person}{Zhuoyan Luo}, \bibinfo{person}{Yong Liu}, \bibinfo{person}{Yue Ma}, \bibinfo{person}{Hengwei Bian}, \bibinfo{person}{Yatai Ji}, \bibinfo{person}{Yujiu Yang}, {and} \bibinfo{person}{Xiu Li}.} \bibinfo{year}{2023}\natexlab{}.
\newblock \showarticletitle{Bridging the Gap: A Unified Video Comprehension Framework for Moment Retrieval and Highlight Detection}.
\newblock \bibinfo{journal}{\emph{arXiv preprint arXiv:2311.16464}} (\bibinfo{year}{2023}).
\newblock


\bibitem[Xiong et~al\mbox{.}(2019)]%
        {xiong2019less}
\bibfield{author}{\bibinfo{person}{Bo Xiong}, \bibinfo{person}{Yannis Kalantidis}, \bibinfo{person}{Deepti Ghadiyaram}, {and} \bibinfo{person}{Kristen Grauman}.} \bibinfo{year}{2019}\natexlab{}.
\newblock \showarticletitle{Less is more: Learning highlight detection from video duration}. In \bibinfo{booktitle}{\emph{Proceedings of the IEEE/CVF conference on computer vision and pattern recognition}}. \bibinfo{pages}{1258--1267}.
\newblock


\bibitem[Xu et~al\mbox{.}(2019)]%
        {xu2019multilevel}
\bibfield{author}{\bibinfo{person}{Huijuan Xu}, \bibinfo{person}{Kun He}, \bibinfo{person}{Bryan~A Plummer}, \bibinfo{person}{Leonid Sigal}, \bibinfo{person}{Stan Sclaroff}, {and} \bibinfo{person}{Kate Saenko}.} \bibinfo{year}{2019}\natexlab{}.
\newblock \showarticletitle{Multilevel language and vision integration for text-to-clip retrieval}. In \bibinfo{booktitle}{\emph{Proceedings of the AAAI Conference on Artificial Intelligence}}, Vol.~\bibinfo{volume}{33}. \bibinfo{pages}{9062--9069}.
\newblock


\bibitem[Xu et~al\mbox{.}(2021)]%
        {xu2021cross}
\bibfield{author}{\bibinfo{person}{Minghao Xu}, \bibinfo{person}{Hang Wang}, \bibinfo{person}{Bingbing Ni}, \bibinfo{person}{Riheng Zhu}, \bibinfo{person}{Zhenbang Sun}, {and} \bibinfo{person}{Changhu Wang}.} \bibinfo{year}{2021}\natexlab{}.
\newblock \showarticletitle{Cross-category video highlight detection via set-based learning}. In \bibinfo{booktitle}{\emph{Proceedings of the IEEE/CVF International Conference on Computer Vision}}. \bibinfo{pages}{7970--7979}.
\newblock


\bibitem[Xu et~al\mbox{.}(2023)]%
        {xu2023mh}
\bibfield{author}{\bibinfo{person}{Yifang Xu}, \bibinfo{person}{Yunzhuo Sun}, \bibinfo{person}{Yang Li}, \bibinfo{person}{Yilei Shi}, \bibinfo{person}{Xiaoxiang Zhu}, {and} \bibinfo{person}{Sidan Du}.} \bibinfo{year}{2023}\natexlab{}.
\newblock \showarticletitle{Mh-detr: Video moment and highlight detection with cross-modal transformer}.
\newblock \bibinfo{journal}{\emph{arXiv preprint arXiv:2305.00355}} (\bibinfo{year}{2023}).
\newblock


\bibitem[Xu et~al\mbox{.}(2024)]%
        {xu2024enhancing}
\bibfield{author}{\bibinfo{person}{Zunnan Xu}, \bibinfo{person}{Jiaqi Huang}, \bibinfo{person}{Ting Liu}, \bibinfo{person}{Yong Liu}, \bibinfo{person}{Haonan Han}, \bibinfo{person}{Kehong Yuan}, {and} \bibinfo{person}{Xiu Li}.} \bibinfo{year}{2024}\natexlab{}.
\newblock \showarticletitle{Enhancing fine-grained multi-modal alignment via adapters: a parameter-efficient training framework for referring image segmentation}. In \bibinfo{booktitle}{\emph{2nd Workshop on Advancing Neural Network Training: Computational Efficiency, Scalability, and Resource Optimization (WANT@ ICML 2024)}}.
\newblock


\bibitem[Xu et~al\mbox{.}(2025)]%
        {xu2025hunyuanportrait}
\bibfield{author}{\bibinfo{person}{Zunnan Xu}, \bibinfo{person}{Zhentao Yu}, \bibinfo{person}{Zixiang Zhou}, \bibinfo{person}{Jun Zhou}, \bibinfo{person}{Xiaoyu Jin}, \bibinfo{person}{Fa-Ting Hong}, \bibinfo{person}{Xiaozhong Ji}, \bibinfo{person}{Junwei Zhu}, \bibinfo{person}{Chengfei Cai}, \bibinfo{person}{Shiyu Tang}, {et~al\mbox{.}}} \bibinfo{year}{2025}\natexlab{}.
\newblock \showarticletitle{Hunyuanportrait: Implicit condition control for enhanced portrait animation}. In \bibinfo{booktitle}{\emph{Proceedings of the Computer Vision and Pattern Recognition Conference}}. \bibinfo{pages}{15909--15919}.
\newblock


\bibitem[Yan et~al\mbox{.}(2023)]%
        {yan2023unloc}
\bibfield{author}{\bibinfo{person}{Shen Yan}, \bibinfo{person}{Xuehan Xiong}, \bibinfo{person}{Arsha Nagrani}, \bibinfo{person}{Anurag Arnab}, \bibinfo{person}{Zhonghao Wang}, \bibinfo{person}{Weina Ge}, \bibinfo{person}{David Ross}, {and} \bibinfo{person}{Cordelia Schmid}.} \bibinfo{year}{2023}\natexlab{}.
\newblock \showarticletitle{Unloc: A unified framework for video localization tasks}. In \bibinfo{booktitle}{\emph{Proceedings of the IEEE/CVF International Conference on Computer Vision}}. \bibinfo{pages}{13623--13633}.
\newblock


\bibitem[Ye et~al\mbox{.}(2021)]%
        {ye2021temporal}
\bibfield{author}{\bibinfo{person}{Qinghao Ye}, \bibinfo{person}{Xiyue Shen}, \bibinfo{person}{Yuan Gao}, \bibinfo{person}{Zirui Wang}, \bibinfo{person}{Qi Bi}, \bibinfo{person}{Ping Li}, {and} \bibinfo{person}{Guang Yang}.} \bibinfo{year}{2021}\natexlab{}.
\newblock \showarticletitle{Temporal cue guided video highlight detection with low-rank audio-visual fusion}. In \bibinfo{booktitle}{\emph{Proceedings of the IEEE/CVF International Conference on Computer Vision}}. \bibinfo{pages}{7950--7959}.
\newblock


\bibitem[Yuan et~al\mbox{.}(2019)]%
        {yuan2019semantic}
\bibfield{author}{\bibinfo{person}{Yitian Yuan}, \bibinfo{person}{Lin Ma}, \bibinfo{person}{Jingwen Wang}, \bibinfo{person}{Wei Liu}, {and} \bibinfo{person}{Wenwu Zhu}.} \bibinfo{year}{2019}\natexlab{}.
\newblock \showarticletitle{Semantic conditioned dynamic modulation for temporal sentence grounding in videos}.
\newblock \bibinfo{journal}{\emph{Advances in Neural Information Processing Systems}}  \bibinfo{volume}{32} (\bibinfo{year}{2019}).
\newblock


\bibitem[Zhang et~al\mbox{.}(2019a)]%
        {zhang2019man}
\bibfield{author}{\bibinfo{person}{Da Zhang}, \bibinfo{person}{Xiyang Dai}, \bibinfo{person}{Xin Wang}, \bibinfo{person}{Yuan-Fang Wang}, {and} \bibinfo{person}{Larry~S Davis}.} \bibinfo{year}{2019}\natexlab{a}.
\newblock \showarticletitle{Man: Moment alignment network for natural language moment retrieval via iterative graph adjustment}. In \bibinfo{booktitle}{\emph{Proceedings of the IEEE/CVF Conference on Computer Vision and Pattern Recognition}}. \bibinfo{pages}{1247--1257}.
\newblock


\bibitem[Zhang et~al\mbox{.}(2024a)]%
        {zhang2024learning}
\bibfield{author}{\bibinfo{person}{Delong Zhang}, \bibinfo{person}{Qiwei Huang}, \bibinfo{person}{Yuanliu Liu}, \bibinfo{person}{Yang Sun}, \bibinfo{person}{Wei-Shi Zheng}, \bibinfo{person}{Pengfei Xiong}, {and} \bibinfo{person}{Wei Zhang}.} \bibinfo{year}{2024}\natexlab{a}.
\newblock \showarticletitle{Learning Implicit Features with Flow Infused Attention for Realistic Virtual Try-On}.
\newblock \bibinfo{journal}{\emph{arXiv preprint arXiv:2412.11435}} (\bibinfo{year}{2024}).
\newblock


\bibitem[Zhang et~al\mbox{.}(2024b)]%
        {zhang2024pixelfade}
\bibfield{author}{\bibinfo{person}{Delong Zhang}, \bibinfo{person}{Yi-Xing Peng}, \bibinfo{person}{Xiao-Ming Wu}, \bibinfo{person}{Ancong Wu}, {and} \bibinfo{person}{Wei-Shi Zheng}.} \bibinfo{year}{2024}\natexlab{b}.
\newblock \showarticletitle{PixelFade: Privacy-preserving Person Re-identification with Noise-guided Progressive Replacement}. In \bibinfo{booktitle}{\emph{Proceedings of the 32nd ACM International Conference on Multimedia}}. \bibinfo{pages}{6326--6334}.
\newblock


\bibitem[Zhang et~al\mbox{.}(2021)]%
        {zhang2021natural}
\bibfield{author}{\bibinfo{person}{Hao Zhang}, \bibinfo{person}{Aixin Sun}, \bibinfo{person}{Wei Jing}, \bibinfo{person}{Liangli Zhen}, \bibinfo{person}{Joey~Tianyi Zhou}, {and} \bibinfo{person}{Rick Siow~Mong Goh}.} \bibinfo{year}{2021}\natexlab{}.
\newblock \showarticletitle{Natural language video localization: A revisit in span-based question answering framework}.
\newblock \bibinfo{journal}{\emph{IEEE transactions on pattern analysis and machine intelligence}} \bibinfo{volume}{44}, \bibinfo{number}{8} (\bibinfo{year}{2021}), \bibinfo{pages}{4252--4266}.
\newblock


\bibitem[Zhang et~al\mbox{.}(2020b)]%
        {zhang2020span}
\bibfield{author}{\bibinfo{person}{Hao Zhang}, \bibinfo{person}{Aixin Sun}, \bibinfo{person}{Wei Jing}, {and} \bibinfo{person}{Joey~Tianyi Zhou}.} \bibinfo{year}{2020}\natexlab{b}.
\newblock \showarticletitle{Span-based localizing network for natural language video localization}.
\newblock \bibinfo{journal}{\emph{arXiv preprint arXiv:2004.13931}} (\bibinfo{year}{2020}).
\newblock


\bibitem[Zhang et~al\mbox{.}(2016)]%
        {zhang2016video}
\bibfield{author}{\bibinfo{person}{Ke Zhang}, \bibinfo{person}{Wei-Lun Chao}, \bibinfo{person}{Fei Sha}, {and} \bibinfo{person}{Kristen Grauman}.} \bibinfo{year}{2016}\natexlab{}.
\newblock \showarticletitle{Video summarization with long short-term memory}. In \bibinfo{booktitle}{\emph{Computer Vision--ECCV 2016: 14th European Conference, Amsterdam, The Netherlands, October 11--14, 2016, Proceedings, Part VII 14}}. Springer, \bibinfo{pages}{766--782}.
\newblock


\bibitem[Zhang et~al\mbox{.}(2020a)]%
        {zhang2020learning}
\bibfield{author}{\bibinfo{person}{Songyang Zhang}, \bibinfo{person}{Houwen Peng}, \bibinfo{person}{Jianlong Fu}, {and} \bibinfo{person}{Jiebo Luo}.} \bibinfo{year}{2020}\natexlab{a}.
\newblock \showarticletitle{Learning 2d temporal adjacent networks for moment localization with natural language}. In \bibinfo{booktitle}{\emph{Proceedings of the AAAI Conference on Artificial Intelligence}}, Vol.~\bibinfo{volume}{34}. \bibinfo{pages}{12870--12877}.
\newblock


\bibitem[Zhang et~al\mbox{.}(2024c)]%
        {zhang2024musetalk}
\bibfield{author}{\bibinfo{person}{Yue Zhang}, \bibinfo{person}{Zhizhou Zhong}, \bibinfo{person}{Minhao Liu}, \bibinfo{person}{Zhaokang Chen}, \bibinfo{person}{Bin Wu}, \bibinfo{person}{Yubin Zeng}, \bibinfo{person}{Chao Zhan}, \bibinfo{person}{Junxin Huang}, \bibinfo{person}{Yingjie He}, {and} \bibinfo{person}{Wenjiang Zhou}.} \bibinfo{year}{2024}\natexlab{c}.
\newblock \showarticletitle{Musetalk: Real-time high quality lip synchronization with latent space inpainting}.
\newblock \bibinfo{journal}{\emph{arXiv preprint arXiv:2410.10122}} (\bibinfo{year}{2024}).
\newblock


\bibitem[Zhang et~al\mbox{.}(2019b)]%
        {zhang2019cross}
\bibfield{author}{\bibinfo{person}{Zhu Zhang}, \bibinfo{person}{Zhijie Lin}, \bibinfo{person}{Zhou Zhao}, {and} \bibinfo{person}{Zhenxin Xiao}.} \bibinfo{year}{2019}\natexlab{b}.
\newblock \showarticletitle{Cross-modal interaction networks for query-based moment retrieval in videos}. In \bibinfo{booktitle}{\emph{Proceedings of the 42nd International ACM SIGIR Conference on Research and Development in Information Retrieval}}. \bibinfo{pages}{655--664}.
\newblock


\end{thebibliography}
\end{document}